\documentclass{article}

    \PassOptionsToPackage{numbers, compress}{natbib}

\usepackage[preprint]{neurips_2026}

\usepackage[utf8]{inputenc} 
\usepackage[T1]{fontenc}    
\usepackage{hyperref}       
\usepackage{url}            
\usepackage{booktabs}       
\usepackage{amsfonts}       
\usepackage{nicefrac}       
\usepackage{microtype}      
\usepackage[table]{xcolor}  

\usepackage{listings}

\lstdefinestyle{promptcode}{
  basicstyle=\ttfamily\footnotesize,
  breaklines=true,
  breakatwhitespace=true,
  columns=fullflexible,
  keepspaces=true,
  frame=single,
  framerule=0.3pt,
  xleftmargin=0.5em,
  xrightmargin=0.5em,
  aboveskip=0.75em,
  belowskip=0.75em,
  showstringspaces=false,
  captionpos=b
}

\makeatletter
\@ifundefined{nolinenumbers}{%
  \newenvironment{nonumberedlisting}{}{}
}{%
  \newenvironment{nonumberedlisting}{\par\begin{nolinenumbers}}{\end{nolinenumbers}\par}
}
\makeatother

\newcommand{\taskmaptitle}[1]{%
  \par\addvspace{0.65\baselineskip}%
  \noindent\textbf{#1}\par\nobreak\vspace{0.25em}%
}

\lstdefinestyle{taskmap}{
  basicstyle=\ttfamily\scriptsize,
  breaklines=true,
  breakatwhitespace=false,
  breakindent=2.2em,
  postbreak=\mbox{\(\hookrightarrow\)\space},
  columns=fullflexible,
  keepspaces=true,
  frame=single,
  framerule=0.3pt,
  framesep=3pt,
  numbers=none,
  showstringspaces=false,
  aboveskip=0.25em,
  belowskip=0.75em,
  xleftmargin=0pt,
  xrightmargin=0pt,
  linewidth=\linewidth
}

\lstdefinestyle{taskmaplong}{
  style=taskmap,
  basicstyle=\ttfamily\tiny,
  breaklines=true,
  breakatwhitespace=false
}

\usepackage{graphicx}

\usepackage{subcaption}

\definecolor{KTLow}{HTML}{F7FBFF}   
\definecolor{KTHigh}{HTML}{08519C}  

\newcommand{\ktcell}[2]{%
  \cellcolor{KTHigh!#1!KTLow}\strut #2%
}

\title{Knowledge-Centric Self-Improvement}

\author{%
  Xuefei (Julie) Wang\\
  Caltech
  \And
  Lauren Hyoseo Yoon\\
  Caltech
  \And
  Chengrui Qu\\
  Caltech
  \And
  Amanda Zichang Wang\\
  Caltech
  \And
  Atharva Sehgal\\
  Caltech
  \And
  Eric Mazumdar\\
  Caltech
  \And
  Yisong Yue\\
  Caltech
}

\usepackage{float}

\usepackage{amsmath}
\usepackage{amssymb}
\usepackage{mathtools}
\usepackage{amsthm}
\usepackage[capitalize,noabbrev]{cleveref}
\usepackage{xspace}
\usepackage{enumitem}
\usepackage{adjustbox}
\usepackage{pifont}
\usepackage{colortbl}
\usepackage{tabularx}
\usepackage{makecell}
\usepackage{multirow}
\usepackage{caption}
\usepackage{array}
\usepackage{wrapfig}
\usepackage{tcolorbox}
\tcbuselibrary{listings,skins}
\usepackage{longtable,ragged2e}

\usepackage{pgfplots}
\pgfplotsset{compat=1.18}

\definecolor{posgreen}{HTML}{E8F3E8}
\definecolor{negred}{HTML}{F8E6E6}
\definecolor{partyellow}{HTML}{FCF4D9}

\usepackage{amsmath,amsfonts,bm}

\def\eqref#1{equation~\ref{#1}}

\def\1{\bm{1}}

\DeclareMathAlphabet{\mathsfit}{\encodingdefault}{\sfdefault}{m}{sl}
\SetMathAlphabet{\mathsfit}{bold}{\encodingdefault}{\sfdefault}{bx}{n}

\usepackage{lineno}

\def\hide#1{\if 0 {#1} \fi}

\DeclareRobustCommand{\xmark}{\ifmmode\text{\ding{55}}\else\ding{55}\fi}
\DeclareRobustCommand{\cmark}{\ifmmode\text{\ding{51}}\else\ding{51}\fi}

\definecolor{darkblue}{rgb}{0, 0, 0.5}
\hypersetup{colorlinks=true, citecolor=darkblue, linkcolor=darkblue, urlcolor=darkblue}

\begin{document}

\maketitle

\begin{abstract}
Self-improving AI systems typically treat the agent as the object that improves, by optimizing prompts, workflows, harnesses, or even the agent’s own code. This agent-centric view can make improvements expensive to maintain and difficult to transfer, because gains become tied to a particular agent design, task distribution, or adaptation run. We study a complementary paradigm: knowledge-centric self-improvement, in which agents remain generic and disposable while the persistent object is a curated knowledge base that agents can leverage for future tasks. We conduct controlled case studies to operationalize this idea via a simple protocol. Agents attempt one task, then contribute evidence-grounded insights to a shared knowledge base via task-level and cross-task forums, followed by knowledge distillation.
Because self-improvement is contained in the knowledge rather than the agent, improvement can be more inspectable, transferable, and portable.
Across abstract reasoning, coding, and terminal benchmarks, this protocol improves solve rates while reducing dollar cost relative to agent-centric baselines. The resulting distilled knowledge also transfers to held-out tasks and across LLM families, indicating that the improvement is not merely an LLM- or run-specific behavior. These results support a new view of self-improving agentic systems: progress can be driven primarily by the curated persistent knowledge. Code is available at \url{https://github.com/recursive-knowledge/KSI}.

\end{abstract}

\section{Introduction}\label{sec:introduction}

The design of self-improving AI systems is rooted in a simple question: when the system gains experience through solving tasks, what should persist and improve over time? 
Most approaches implicitly answer this question by making the agent the persistent object of improvement, through prompt and workflow optimization \citep{fernandoPromptbreederSelfreferentialSelfimprovement2024,zhangAgenticContextEngineering2026,zhugeGPTSwarmLanguageAgents2024}, harness search \citep{metaharness}, or self-modification of the agent's underlying code \citep{dgm,hgm,hu2025automateddesignagenticsystems,yinGodelAgentSelfReferential2025,xiaLiveSWEagentCanSoftware2025}. This agent-centric view has produced strong systems, but it also makes improvement expensive to maintain. A single persistent agent must absorb many local lessons, some of which are task-specific, redundant, or mutually inconsistent; as the agent grows, useful behavior can be diluted by conflicting updates, and each new adaptation might degrade performance on the previous tasks.

\begin{figure}[t]
\centering
\includegraphics[width=0.95\textwidth]{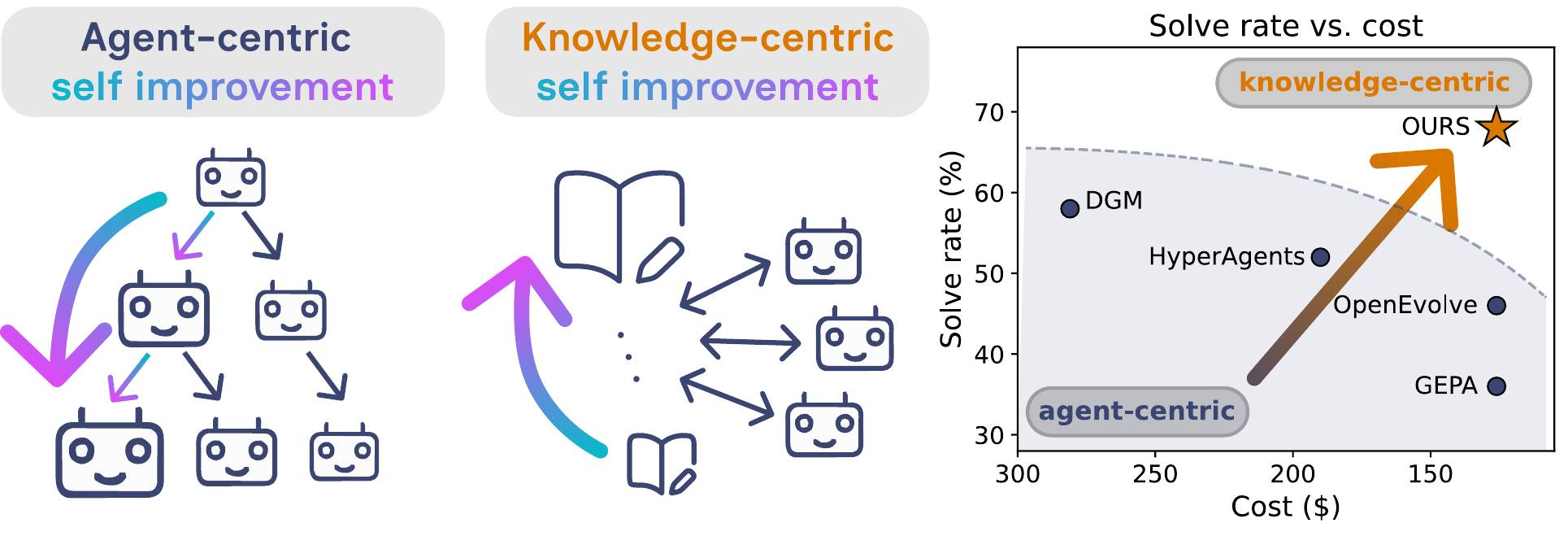}

\caption{\textbf{From agent-centric to knowledge-centric self-improvement.} \textit{(Left)} Conventional self-improving agentic systems treat the agent as the persistent substrate, evolving prompts, workflows, or agent code. \textit{(Middle)} We invert this design. Agents are kept generic and disposable, while the persistent improving substrate is a shared knowledge base, curated through a simple protocol to generate and utilize distilled insights. \textit{(Right)} Across our benchmarks, shifting from agent-centric to knowledge-centric self-improvement pushes the Pareto frontier forward.}
\label{fig:main}
\vspace{-2ex}
\end{figure}

We study a complementary paradigm in which the persistent object is a curated knowledge base rather than the agent itself (Figure~\ref{fig:main}). In this \emph{knowledge-centric self-improvement} paradigm, agents are transient workers that read from and write to the shared knowledge. The knowledge base accumulates task evidence, uses forum discussion to compare and reconcile seemingly conflicting opinions, and exposes distilled guidance to future agents. Compared to agent-centric approaches, this shift can make self-improvement cheaper to maintain and easier to reuse: the system no longer has to preserve an increasingly specialized agent, but instead maintains an external knowledge base that can be inspected, condensed, and shared across fresh agents. Rather than coupling each improvement to a particular agent design or adaptation run, the system extracts reusable knowledge that can transfer across tasks and LLM families.

To isolate persistent curated knowledge as the improvement mechanism, we keep agents generic, stateless, and disposable. Each agent starts with a fresh context, receives the relevant distilled knowledge bundle, attempts one task with standard tool use, and contributes its experience back to the shared knowledge base. Agents do not carry persistent private memory, role specialization, task-specific architecture, or custom orchestration. The only object that changes is the curated knowledge base.

We instantiate this design with a deliberately simple three-stage curation protocol, organized as threaded discussions in which agents post claims, cite one another, and reply with supporting or challenging evidence. After task execution, each agent posts to a task-level forum, turning its attempt into evidence-grounded local claims about what worked, what failed, and which constraints or hypotheses mattered, which peer agents then support or challenge. A cross-task forum then brings these claims into open discussion across tasks: agents respond to peers' posts, and recurring principles, disagreements, and failure modes surface through this exchange. Finally, a distillation stage consolidates the surviving local and global claims into compact bundles that future fresh agents consume. We do not claim that this protocol is optimal; rather, it provides a controlled way to test whether curated knowledge alone can drive self-improvement when agent design is held fixed.

This paradigm goes beyond standard experience reuse or memory architectures \citep{memgpt,memorybank,gmemory,zhao2024expelllmagentsexperiential,wangVoyagerOpenEndedEmbodied2023,wang2024agentworkflowmemory,tang2025agentkbleveragingcrossdomain} in what is stored and retrieved for future agents before they act. Those systems accumulate the agents' own successes and failures, whereas our knowledge curation protocol converts experience into deeper, more scoped, evidence-grounded guidance.
Therefore, we treat disagreement as useful evidence rather than instability. Task-level forums preserve competing local hypotheses, cross-task forums test which ones recur or fail elsewhere, and distillation keeps only the claims that remain actionable and supported. Moreover, prior systems still rely on persistent or specialized solving agents. In contrast, our agents are generic and disposable — the curated knowledge base is the only object that changes across tasks.

We evaluate on three categories of tasks, namely abstract reasoning (ARC-AGI-1 \citep{arcagi1}, ARC-AGI-2 \citep{arcagi2}), coding (Polyglot \citep{polyglot}, SWE-bench Pro \citep{swebenchpro}), and terminal skills (Terminal-Bench 2 \citep{terminalbench2}). Our contributions and findings are as follows:
\begin{itemize} [leftmargin=12pt]
    \item \textbf{We develop a controlled framework for studying knowledge-centric self-improvement.}  We introduce a protocol in which the only improving object is a shared knowledge base, while agents are fixed as generic and disposable. This setup isolates curated knowledge as the source of improvement, and provides a testbed for studying knowledge-driven gains independently of agent design.
    \item \textbf{Knowledge curation outperforms agent-centric optimization with higher cost efficiency.} 
    Compared against strong agent-centric self-improvement baselines, including Darwin G\"odel Machine (DGM) \citep{dgm}, HyperAgents \citep{hyperagents}, and Meta-Harness \citep{metaharness}, generic agents paired with a curated knowledge base achieve higher solve rates across the evaluated benchmarks, often at substantially lower dollar cost.
    \item \textbf{Curated knowledge transfers across tasks and across LLM families.} Distilled bundles curated on a training set improve zero-shot performance on a held-out test set, and bundles curated under one LLM family remain effective when consumed by another. This is evidence that curation yields an artifact whose utility extends beyond the run that generated it.
\end{itemize}
Taken together, these results highlight the advantage of curating an efficiently reusable knowledge artifact, rather than having improvements be coupled to the agent design.
\section{Related Work}\label{sec:related-work}

We distinguish prior work along two axes: whether the persistent object is the agent, a memory store, or an external knowledge base; and whether past experience is stored, retrieved, optimized, or curated into transferable guidance.

\textbf{Self-improving systems.}
A large body of work on self-improving AI systems treats the agent as the object that improves over time. Iterative gains are achieved by modifying the agent's prompts \citep{assumpcaoCodeEvolveOpenSource2026,fernandoPromptbreederSelfreferentialSelfimprovement2024,wangEvolvingPromptsInContext2025,yeMetaContextEngineering2026,zelikmanSTaRBootstrappingReasoning2022,zhangAgenticContextEngineering2026}, workflows \citep{zhugeGPTSwarmLanguageAgents2024}, learned skill and tool libraries \citep{wuOSCopilotGeneralistComputer2024}, model weights \citep{akibaEvolutionaryOptimizationModel2025,zhangNatureInspiredPopulationBasedEvolution2025}, harness search \citep{metaharness}, or even the agent's underlying codebase \citep{hu2025automateddesignagenticsystems,hgm,dgm,hyperagents,xiaLiveSWEagentCanSoftware2025,yinGodelAgentSelfReferential2025}. Despite their differences, these methods share a common premise: self-improvement is achieved by making the agent more capable through better self-feedback, stronger internal structure, or more sophisticated self-modification. Our work studies a different design dimension: we hold the agent fixed and ask whether improvement can compound when the only changing object is a curated knowledge base.

\textbf{Agent memory architectures and benchmarks.}
A parallel line of work develops memory architectures that let agents persist and organize information across interactions. Building on retrieval augmentation \citep{rag,retro}, many systems study how to store, retrieve, link, and maintain memories so that agents can better leverage prior context \citep{memgpt,memorybank,xu2025amemagenticmemoryllm,fang2026mempexploringagentprocedural}. Recent work further improves the memory substrate itself, moving from flat retrieval toward structured, production-ready, or self-evolving memory systems \citep{chhikara2025mem0buildingproductionreadyai,zhang2026memrlselfevolvingagentsruntime,zhangMemEvolveMetaEvolutionAgent2025}. Complementary benchmarks such as LoCoMo and LongMemEval evaluate whether agents can retain and reason over information across long conversational histories \citep{maharana2024locomo,wu2024longmemeval}. These works are largely organized around the continuity problem: how an agent should retain, update, and retrieve useful context over long interactions, episodes, or task histories. We instead treat memory as storage infrastructure rather than the object of study, and make evidence-grounded knowledge curation the mechanism of improvement. Our central question is not how to remember more, but how raw attempts become scoped, transferable guidance that improves over generations while agents remain disposable.

\textbf{Experience reuse and knowledge transfer.}
The closest prior work externalizes past experience into reusable artifacts: executable skills or workflows
\citep{wangVoyagerOpenEndedEmbodied2023,wang2024agentworkflowmemory,zheng2025skillweaverwebagentsselfimprove},
retrieved lessons or guidelines
\citep{zhao2024expelllmagentsexperiential,fu2024autoguideautomatedgenerationselection},
and higher-level templates, principles, or structured knowledge bases
\citep{yang2024bufferthoughtsthoughtaugmentedreasoning,wu2026evolverselfevolvingllmagents,puPiFlowPrincipleAwareScientific2026,gmemory,tang2025agentkbleveragingcrossdomain}.
The core difference between our approach and prior work lies in how experience is transformed into knowledge. Prior work relies on agent-side abstraction: single agents compress their own trajectories into reusable artifacts. We instead treat knowledge as evidence adjudicated through discussion among multiple attempts and multiple agents.
Each attempt contributes claims to a task-level forum of competing hypotheses, where peers reply with supporting or contradicting evidence; the surviving claims are then tested for recurrence across tasks, and a distillation step retains only the principles that apply to new tasks. This shifts the burden of reliability from the agent to the evidentiary process.
Therefore, the agents themselves can remain simple and disposable, leaving the shared knowledge base as the only evolving component. We then test whether this knowledge successfully transfers to new tasks and different LLM families.
\section{Knowledge Curation Protocol}
\label{sec:methodology}

Our protocol turns each (disposable) agent's task attempt into evidence that the next generation of agents inherits, while the agents themselves do not change across generations. Every agent is a fresh instance (i.e., with a clean context) that reads from the shared knowledge base, attempts a task, and writes evidence back to it. All persistent state lives in the knowledge base, which is repeatedly curated into a form that later agents can use.

\subsection{System Overview}
\label{subsec:overview}

The framework is designed to isolate knowledge curation as the only mechanism of self-improvement, to enable controlled experimentation. Agents are transient  instances: each receives a task description, standard tool access, and a seed bundle drawn from the shared knowledge base, attempts the task in a single solving session, and records its outcome. Benchmark-specific in-session submission budgets (e.g., ARC's two blind trials per test input) follow the official benchmark protocols and are listed in Appendix~\ref{app:hyperparameters}. The agents do not inherit private memory, identity, specialization, or modified prompts from previous agents. All persistent state lives in the knowledge base.

The knowledge base stores three artifacts: a typed attempt table, forum posts, and distilled bundles. The attempt table records each attempt and outcome. Forum posts turn execution traces into evidence-grounded claims about why an attempt worked, failed, or should be modified. Distilled bundles compress the surviving claims into reusable guidance for future agents.

\begin{wrapfigure}{r}{0.4\textwidth}
\vspace{-3ex}
\centering
\includegraphics[width=0.35\textwidth]{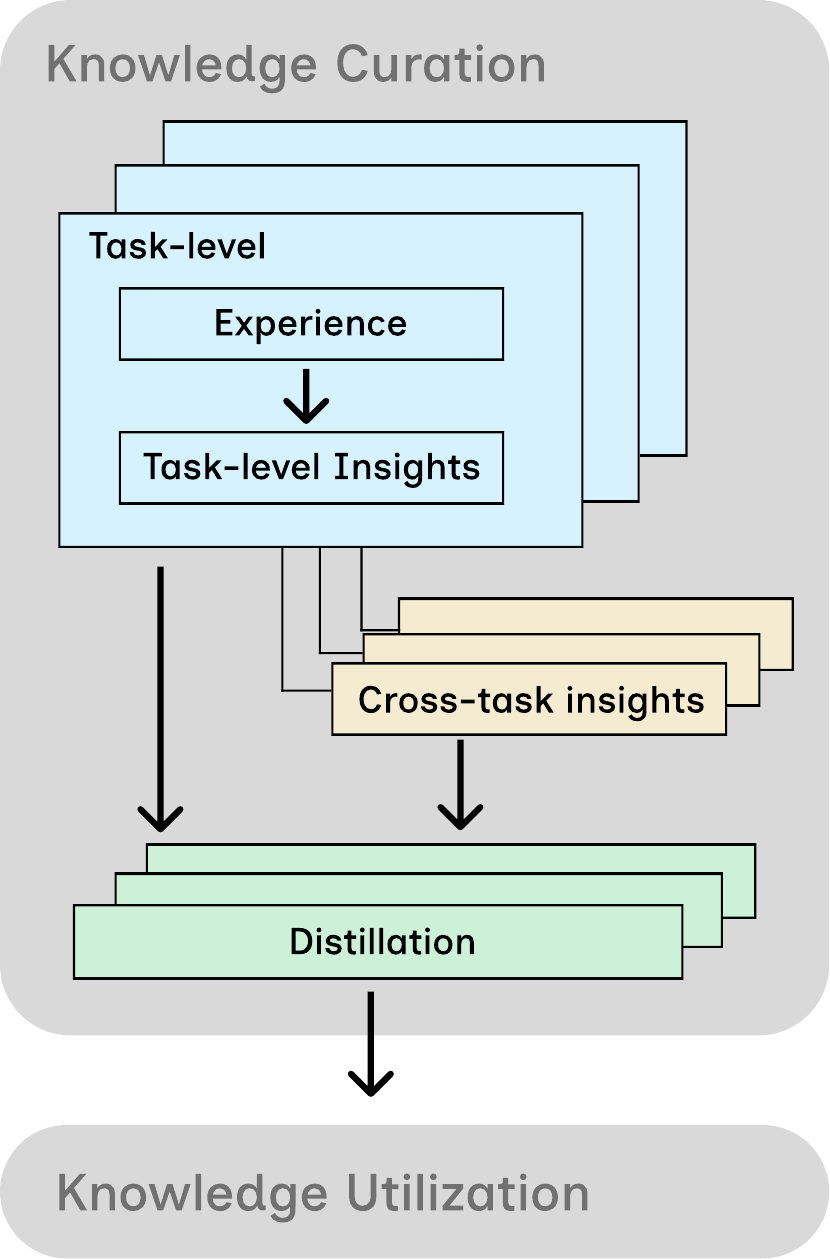}
\vspace{1ex}
\caption{Knowledge curation protocol consisting of three stages. \textbf{Task-level forum} (agents propose task-level posts summarizing local outcomes and evidence on the current task), \textbf{Cross-task forum} (agents discuss transferable patterns and review peer posts with supporting or challenging evidence), and \textbf{Distillation} (forum posts are consolidated into typed task-level and cross-task bundles written back to the shared knowledge base). The schemas of the task-level posts, cross-task posts, and distilled bundles are listed in Appendix~\ref{app:protocol-schemas}.}
\vspace{-3ex}
\label{fig:knowledge_curation_schema}
\end{wrapfigure}

Once agents complete their attempts, the system curates the knowledge base through three stages, as depicted in Figure~\ref{fig:knowledge_curation_schema}.
The two forum stages take the familiar form of an online discussion board: each task has its own thread, a shared thread spans all tasks in the generation, and agents post evidence-grounded claims, cite earlier posts by id, and reply with supporting or challenging evidence.
The three stages form an abstraction ladder. Task-level forums preserve local evidence and discussion from individual attempts, cross-task forums test which claims survive debate across tasks, and distillation converts the surviving claims into compact seed bundles.

\begin{figure}[h!]
    \centering
    \begin{subfigure}{\linewidth}
        \centering
        \includegraphics[width=1.0\linewidth]{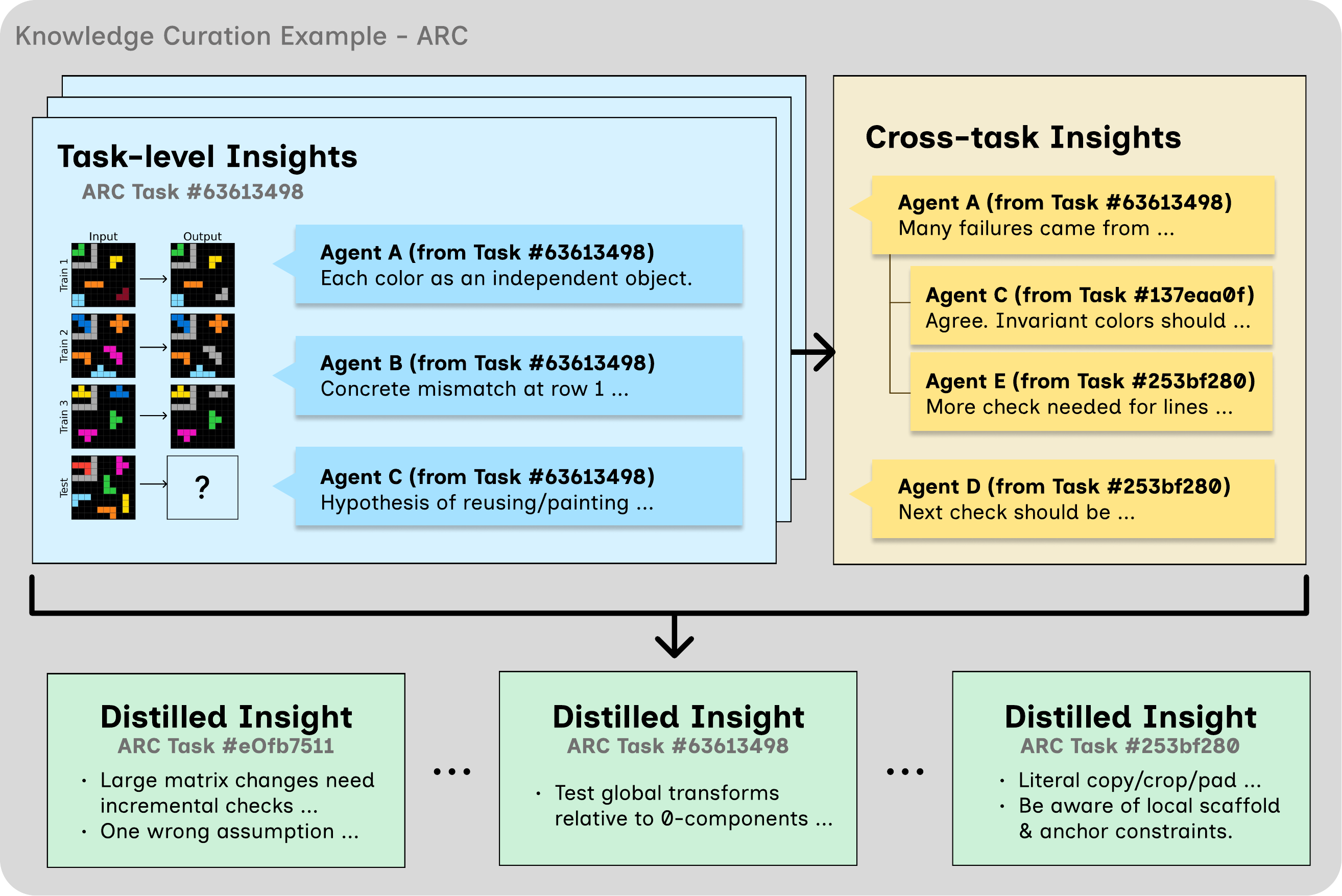}
        \caption{Example of the knowledge curation pipeline applied to an Abstract Reasoning Corpus (ARC) task.}
        \label{fig:arc_knowledge_ex}
    \end{subfigure}

    \vspace{10pt} 

    \begin{subfigure}{\linewidth}
        \centering
        \includegraphics[width=1.0\linewidth]{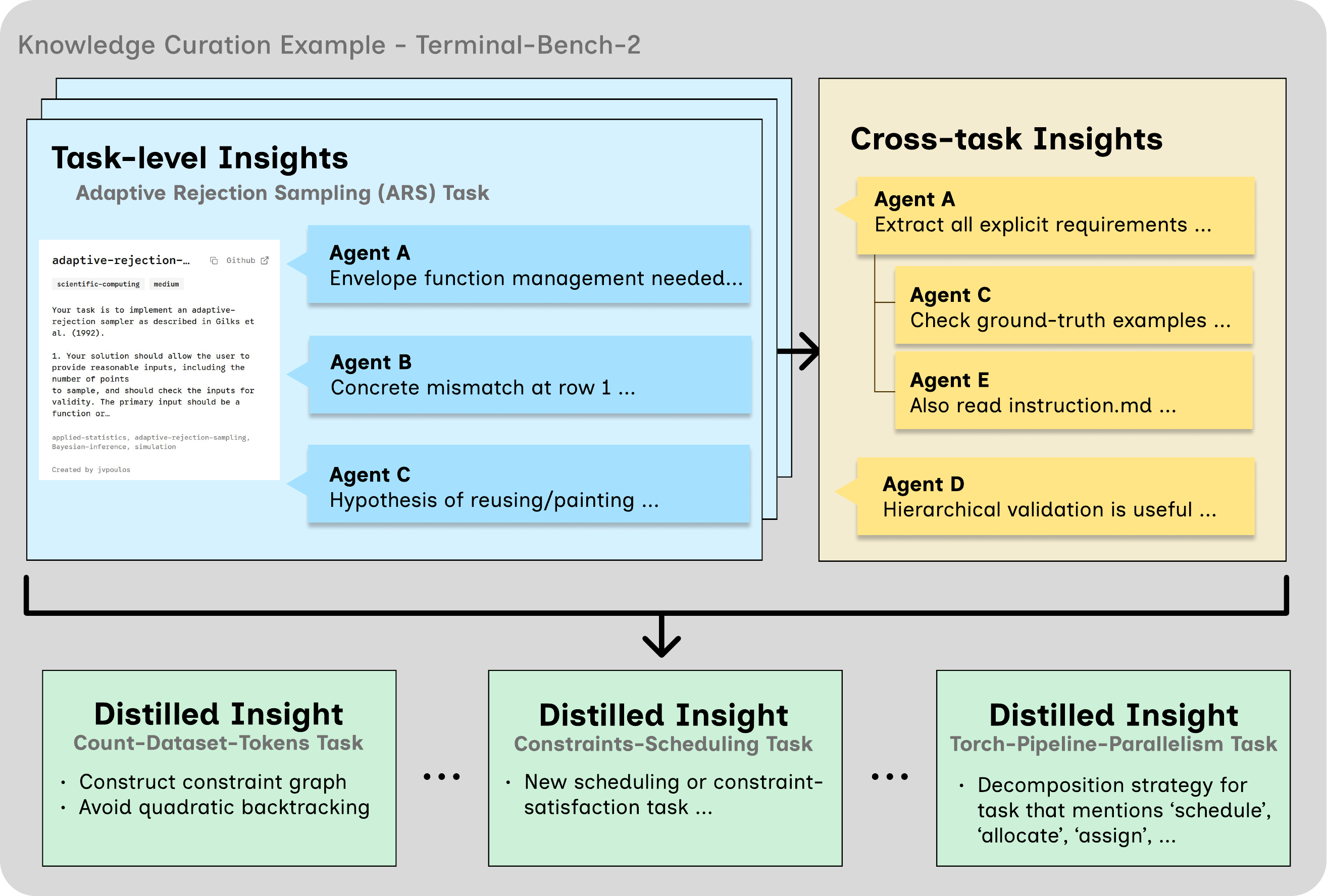}
        \caption{Example of the knowledge curation pipeline applied to a Terminal-Bench 2 task.}
        \label{fig:tb2_knowledge_ex}
    \end{subfigure}
    \caption{Examples of knowledge curation on two task families. Local task evidence is first organized into task-specific guidance, then compared against cross-task patterns, and finally distilled into seed bundles that future agents can use. Detailed renderings of both examples are in Appendix~\ref{app:knowledge-examples}.
    }
    \label{fig:combined_knowledge}
\end{figure}

\textbf{Stage 1: Task-level forum.}
The task-level forum converts execution traces into task-specific insights through discussion on a per-task thread. After attempting a task, each agent posts to that task's thread, recording not only its final answer or patch, but also the hypotheses it tested, constraints it confirmed, checks that were informative, and strategies that led to failure. Posting is not write-only: agents read the earlier posts in the thread and cite the specific posts that support or contradict their own experience, so successive posts refine, challenge, or rule out one another's hypotheses. Forum posts are grounded in the local task instance: for ARC tasks, this might include invariant colors, object boundaries, rejected transformations, or concrete mismatch locations against the training examples; for coding and terminal tasks, it might include API signatures, test-runner behavior, file-path assumptions, dependency issues, or edge cases exposed by failed tests (Figure~\ref{fig:combined_knowledge}). The output of this stage is local guidance for the same task: what appears to matter, what should not be retried, and what a later fresh agent should verify before acting.

\textbf{Stage 2: Cross-task forum.}
The cross-task forum decides which local observations should survive beyond the task that produced them. Agents discuss task-level posts from the current generation and propose claims that recur across tasks, such as common error types, verification strategies, invariants, environment assumptions, or decomposition patterns. To prevent generic advice from accumulating, each cross-task claim must be grounded in concrete evidence from one or more attempts, and later posts take an explicit stance toward prior claims: agree, disagree, or synthesize. This stage treats disagreement as evidence rather than instability. Conflicting local claims expose which guidance is task-specific, which failure modes recur, and which principles remain useful under contrary evidence (Figure~\ref{fig:combined_knowledge}); Appendix~\ref{app:disagreement-examples} traces recorded examples of this mechanism end-to-end, from conflicting forum posts through scoped distilled claims to later-generation solves.
The output is ideally a set of transferable claims: grounded guidance that is no longer tied to a single task attempt. Section~\ref{sec:experiments.transfer} evaluates whether the cross-task knowledge transfers to unseen tasks.

\textbf{Stage 3: Distillation.}
The third stage turns the accumulated evidence into the artifact consumed by future agents. Distillation produces per-task bundles and a cross-task bundle. Both bundle types use the same typed fields, namely transferable insights, confirmed constraints, rejected hypotheses, pitfalls, checks, and next steps.
Distillation is therefore designed as a selection step that is tailored to the new task rather than a generic summarization step: the distiller LLM is instructed to keep claims that are actionable, evidence-grounded, and scoped, and to drop vague advice that does not name the condition under which it applies. The next generation receives the distilled bundles before they act (Figure~\ref{fig:combined_knowledge}).
The protocol makes knowledge, not agent state, the unit that improves. Because agents are re-instantiated for each attempt, any performance gain across generations must come from the accumulated and distilled knowledge they receive.
This design enables controlled evaluation of two important questions. First, whether curated knowledge increases task-solving performance; and second, whether the resulting artifact transfers beyond the tasks and LLM family that produced it. We describe the experimental setup and findings in Section~\ref{sec:experiments}.

\section{Experiments}\label{sec:experiments}

We evaluate whether curated knowledge can drive self-improvement across real-world coding, terminal-execution, and abstract-reasoning tasks. These settings reflect the intended use of our protocol, where agents repeatedly solve heterogeneous tasks under practical performance and efficiency constraints.
We compare against the two predominant families of LLM-guided self-improvement: agent-centric systems \citep{hyperagents,dgm,metaharness} and prompt optimization frameworks \citep{openevolve,agrawal2025gepareflectivepromptevolution}.
Concretely, our experiments address four questions:

\begin{itemize}[leftmargin=*, itemsep=0pt, topsep=0pt, parsep=0pt]
\item \textbf{Does knowledge curation outperform agent-centric self-improvement?} (\S~\ref{sec:experiments.agents})

\item \textbf{Does it outperform prompt optimization?} (\S~\ref{sec:experiments.prompts})

\item \textbf{Does the same protocol remain effective across different LLM families?} (\S~\ref{sec:experiments.llms})

\item \textbf{Does the curated knowledge transfer to held-out tasks and different LLMs?} (\S~\ref{sec:experiments.transfer})

\end{itemize}

\subsection{Agent-Centric Baselines}
\label{sec:experiments.agents}

\textbf{Benchmarks and protocol.}
We evaluate on five benchmarks spanning coding, terminal skills, and abstract reasoning. A more in-depth discussion on benchmarks is presented in Appendix~\ref{app:benchmark-details}. Polyglot~\citep{polyglot} evaluates code generation and editing across C++, Go, Java, JavaScript, Python, and Rust, and we use the 50-task subset adopted by Darwin G\"odel Machine (DGM)~\citep{dgm} and HyperAgents~\citep{hyperagents}. SWE-bench Pro~\citep{swebenchpro} evaluates realistic software-engineering agents on harder, less contaminated, and more diverse repository tasks, and from the 731 public instances we sample a 50-task subset uniformly at random with a fixed seed. ARC-AGI-1 and ARC-AGI-2~\citep{arcagi1,arcagi2} evaluate abstract visual reasoning by requiring exact-grid reconstruction from a few input-output examples, and we sample 50 tasks from each benchmark using the official exact-match scoring protocol. Terminal-Bench 2~\citep{terminalbench2} evaluates agents in 89 real terminal environments with containerized tasks, human-written reference solutions, and verification tests.

In principle, agents can run asynchronously and in a decentralized manner. For practicality and direct comparability with agent-centric baselines, we implement a generation-based system. We begin with a pool of 50 tasks for each coding and abstract-reasoning benchmark. At the start of each generation, we spawn one agent per unsolved task. Each agent receives the typed attempt table for its assigned task, the per-task distilled insights for that task, and the cross-task distilled insights from the global knowledge base. After execution, agents contribute back through the three-stage curation protocol of Section~\ref{sec:methodology}. Solved tasks are removed from the active pool, so the system does not spend tokens revisiting tasks already passed. We run 10 generations per benchmark and for our approach we run 3 seeds.

\textbf{Agents and baselines.}
Task-solving agents are implemented as off-the-shelf calls to the Anthropic Claude Agent SDK with Haiku 4.5 and the OpenAI Agents SDK with GPT-5.4-mini using medium reasoning; the forum and distillation phases of the curation protocol instead call the same LLMs through direct API loops without agent tooling. For coding and terminal benchmarks, agents receive the standard SDK tool set of file read/write, shell execution, and test invocation. Our system and the agent-centric baselines draw from this same tool set.

We compare against two representative agent-centric self-improving baselines, Darwin G\"odel Machine (DGM)~\citep{dgm} and HyperAgents~\citep{hyperagents}. Both frameworks improve performance through iterative refinement of agent behavior, prompts, strategies, or reasoning processes. For Terminal-Bench 2, we compare directly against strong reported systems including OpenHands~\citep{wang2025openhandsopenplatformai}, Terminus 2~\citep{terminalbench2}, Mini-SWE-Agent~\citep{yang2024sweagent}, Terminus-KIRA~\citep{terminuskira}, Goose~\citep{goose}, and Meta-Harness~\citep{metaharness}. All comparator scores are the Haiku~4.5 entries compiled in Table~7 of the Meta-Harness paper~\citep{metaharness}. Every listed system uses a Haiku~4.5 base agent; the Meta-Harness score combines Claude Opus 4.6 as proposer with Haiku 4.5 as the base agent. These baselines improve the agent, execution scaffold, or orchestration loop, while our method instead keeps agents simple and externalizes improvement into a shared curated knowledge base.

\begin{table*}[t]
\centering
\small

\begin{subtable}{\textwidth}
\centering
\caption{Abstract Reasoning \emph{(left)} and Coding \emph{(right)} Benchmarks.}
\label{tab:baseline_reasoning_coding}
\footnotesize
\setlength{\tabcolsep}{3pt}
\begin{tabular}{l cc cc | cc cc}
\toprule
& \multicolumn{4}{c|}{\textbf{Abstract Reasoning}} & \multicolumn{4}{c}{\textbf{Coding}} \\
\cmidrule(lr){2-5} \cmidrule(lr){6-9}
& \multicolumn{2}{c}{ARC-AGI-1} & \multicolumn{2}{c|}{ARC-AGI-2} & \multicolumn{2}{c}{Polyglot} & \multicolumn{2}{c}{SWE-bench Pro} \\
\cmidrule(lr){2-3} \cmidrule(lr){4-5} \cmidrule(lr){6-7} \cmidrule(lr){8-9}
Method & Solve $\uparrow$ & Cost $\downarrow$ & Solve $\uparrow$ & Cost $\downarrow$ & Solve $\uparrow$ & Cost $\downarrow$ & Solve $\uparrow$ & Cost $\downarrow$ \\
\midrule
OURS$_{\mathrm{haiku\,4.5}}$ & \textbf{86.7\%{\scriptsize$\pm$4.2}} & \textbf{\$76{\scriptsize$\pm$16}} & \textbf{82.7\%{\scriptsize$\pm$6.1}} & \textbf{\$80{\scriptsize$\pm$1}} & \textbf{68.0\%{\scriptsize$\pm$2.0}} & \textbf{\$126{\scriptsize$\pm$6}} & \textbf{64.0\%{\scriptsize$\pm$2.0}} & \textbf{\$208{\scriptsize$\pm$19}} \\
\midrule
HyperAgents$_{\mathrm{haiku\,4.5}}$ \citep{hyperagents} & 70\% & \$234 & 60\% & \$188 & 52\% & \$190 & 42\% & \$431 \\
DGM$_{\mathrm{haiku\,4.5}}$ \citep{dgm} & n/a & n/a & n/a & n/a & 58\% & \$281 & 54\% & \$713 \\
\bottomrule
\end{tabular}
\end{subtable}

\vspace{1.5em}

\begin{subtable}[c]{0.45\textwidth}
\centering
\caption{Terminal Skill Benchmark.}
\label{tab:baseline_terminal}
\begin{tabular}{l|c}
\toprule
Method & Terminal-Bench 2 \\
\midrule
OpenHands \citep{wang2025openhandsopenplatformai} & 13.9\% \\
Terminus 2 \citep{terminalbench2} & 28.3\% \\
 Mini-SWE-Agent \citep{yang2024sweagent} & 29.8\% \\
Terminus-KIRA \citep{terminuskira} & 33.7\% \\
Goose \citep{goose} & 35.5\% \\
Meta-Harness \citep{metaharness} & 37.6\% \\
\midrule
OURS & \textbf{43.8\%{\scriptsize$\pm$3.4}} \\
\bottomrule
\end{tabular}
\end{subtable}%
\hfill%
\begin{minipage}[c]{0.52\textwidth}
\caption{\small Baseline comparisons across task families using Haiku 4.5 as the LLM. \textbf{(a)} Performance on abstract reasoning and coding benchmarks over 10 generations, where the two task families share a common row header. OURS entries are mean{\scriptsize$\pm$}standard deviation over three seeds; baselines are single runs. All baselines are rerun under our evaluation protocol. DGM is evaluated only on coding benchmarks (SWE-bench and Polyglot) in the original work and self-modifies its ``coding capabilities''~\citep{dgm}, so we mark it n/a on ARC-AGI-1/2. \textbf{(b)} Performance on Terminal-Bench 2. Baseline scores are the Haiku~4.5 entries compiled in Table~7 of the Meta-Harness paper~\citep{metaharness}; note that the Meta-Harness score combines a Claude Opus 4.6 proposer with a Haiku 4.5 base agent. }
\label{tab:agent_baselines}
\end{minipage}

\vspace{-1em}
\end{table*}

\textbf{Results.}
Table~\ref{tab:agent_baselines} shows that agents using our knowledge curation protocol achieve the highest solve rates among the listed Haiku-based methods on ARC-AGI-1, ARC-AGI-2, Polyglot, and SWE-bench Pro, while also using the lowest reported costs. On Terminal-Bench 2, one of the hardest agentic benchmarks, our simple agent framework equipped with knowledge-centric improvement achieves competitive performance against multiple strong agentic coding systems. The qualitative Terminal-Bench examples in Figures~\ref{fig:tb2_knowledge_ex} and~\ref{fig:tb2_full_knowledge_ex} (see Figures~\ref{fig:arc_knowledge_ex} and~\ref{fig:arc_full_knowledge_ex} for ARC-AGI counterparts) illustrate the mechanism behind these gains: later iterations reuse distilled insights and failure signatures about environment assumptions, missing artifacts, and incorrect workflows, rather than repeatedly rediscovering the same bottlenecks.

\subsection{Prompt Optimization}
\label{sec:experiments.prompts}

A second well-studied form of LLM self-improvement optimizes a reusable solver prompt instead of the agent or its surrounding artifacts.

\textbf{Setup.}
Prompt optimization targets a different object of improvement from our method. We therefore compare against GEPA~\citep{agrawal2025gepareflectivepromptevolution} and OpenEvolve~\citep{openevolve} on ARC-AGI-1 and Polyglot, which provide representative abstract-reasoning and code-reasoning settings. GEPA and OpenEvolve optimize a reusable solver or coding-policy prompt over the same fixed train-50 task pool. We match the improvement budget to the realized cost of our main experiments on each benchmark (Table~\ref{tab:agent_baselines}). Under these budgets, GEPA runs for 15 iterations on ARC-AGI-1 and 7 iterations on Polyglot, while OpenEvolve runs for 81 iterations on ARC-AGI-1 and 46 iterations on Polyglot.

\begin{wraptable}{r}{0.5\textwidth}
\centering
\small
\begin{tabular}{l|cc}
\toprule
Method & ARC-AGI-1 & Polyglot \\
\midrule
OURS & \textbf{86.7\%{\scriptsize$\pm$4.2}} & \textbf{68.0\%{\scriptsize$\pm$2.0}} \\
\midrule
GEPA \citep{agrawal2025gepareflectivepromptevolution} & 44\% & 36\% \\
OpenEvolve \citep{openevolve} & 54\% & 46\% \\
\bottomrule
\end{tabular}
\vspace{-0.1em}
\caption{Comparison against Prompt Optimization Methods. We report the best score achieved by running iterations that match the budget of our main experiments. OURS is mean{\scriptsize$\pm$}standard deviation over three seeds. As these prompt optimization methods showed limited gains on ARC-AGI-1 and Polyglot, we did not extend these baselines to more resource-intensive benchmarks.}
\label{tab:baseline_prompt}
\vspace{-2em}
\end{wraptable}
\textbf{Results.}
Table~\ref{tab:baseline_prompt} shows that knowledge curation outperforms both prompt-optimization baselines under the matched budgets. This comparison isolates a distinct self-improvement axis: prompt optimizers refine a reusable policy prompt, whereas our method refines a reusable knowledge artifact that records distilled insights, environment assumptions, and recurring failure modes. As these prompt optimization-oriented methods showed limited gains on ARC-AGI-1 and Polyglot, we did not extend these baselines to the more resource-intensive benchmarks for cost budget reasons.

\subsection{LLM Generalization}
\label{sec:experiments.llms}

The previous experiments fix the LLM and vary the comparison method. We now reverse this and ask whether the same procedure remains effective when the LLM family changes.

\textbf{Setup.}
To test whether the protocol is tied to a single LLM family, we run the same knowledge curation procedure with Haiku 4.5 and GPT-5.4-mini (medium reasoning) across the coding and abstract-reasoning benchmarks. Each run uses the same 50-task pools and 10-generation budget described above. LLMs, hyperparameters, and task maps follow Appendix~\ref{app:hyperparameters} and Appendix~\ref{app:task-maps}.

\textbf{Results.}
Table~\ref{tab:cross_model} shows that the protocol remains effective under both LLMs. GPT-5.4-mini attains the higher mean solve rate and lower cost on all four benchmarks, while the Haiku 4.5 runs still exceed every agent-centric baseline in Table~\ref{tab:agent_baselines} at a fraction of the cost. These results suggest that the curated knowledge protocol is not tied to a single LLM family, although the margins depend on the LLM.

\begin{table}[H]

\centering
\small
\setlength{\tabcolsep}{3pt}
\begin{tabular}{l cc cc | cc cc}
\toprule
& \multicolumn{4}{c|}{\textbf{Coding}} & \multicolumn{4}{c}{\textbf{Abstract Reasoning}} \\
\cmidrule(lr){2-5} \cmidrule(lr){6-9}
& \multicolumn{2}{c}{Polyglot} & \multicolumn{2}{c|}{SWE-bench Pro} & \multicolumn{2}{c}{ARC-AGI-1} & \multicolumn{2}{c}{ARC-AGI-2} \\
\cmidrule(lr){2-3} \cmidrule(lr){4-5} \cmidrule(lr){6-7} \cmidrule(lr){8-9}
Method & Solve $\uparrow$ & Cost $\downarrow$ & Solve $\uparrow$ & Cost $\downarrow$ & Solve $\uparrow$ & Cost $\downarrow$ & Solve $\uparrow$ & Cost $\downarrow$ \\
\midrule
OURS$_{\mathrm{haiku\,4.5}}$ & 68.0\%{\scriptsize$\pm$2.0} & \$126{\scriptsize$\pm$6} & 64.0\%{\scriptsize$\pm$2.0} & \$208{\scriptsize$\pm$19} & 86.7\%{\scriptsize$\pm$4.2} & \$76{\scriptsize$\pm$16} & 82.7\%{\scriptsize$\pm$6.1} & \$80{\scriptsize$\pm$1} \\
OURS$_{\mathrm{gpt\,5.4-mini}}$ & \textbf{72.7\%{\scriptsize$\pm$2.3}} & \textbf{\$56{\scriptsize$\pm$1}} & \textbf{70.7\%{\scriptsize$\pm$2.3}} & \textbf{\$157{\scriptsize$\pm$11}} & \textbf{93.3\%{\scriptsize$\pm$7.0}} & \textbf{\$16{\scriptsize$\pm$3}} & \textbf{90.0\%{\scriptsize$\pm$5.3}} & \textbf{\$20{\scriptsize$\pm$1}} \\
\bottomrule
\end{tabular}
\vspace{0.5em}
\caption{Cross-LLM Generalization. This table demonstrates the generalization and cost efficiency of our framework across different LLM families. GPT-5.4-mini runs use medium reasoning effort (Section~\ref{sec:experiments.agents}).}
\label{tab:cross_model}
\end{table}

\subsection{Held-Out Knowledge Transfer}
\label{sec:experiments.transfer}

The curated knowledge base is itself an output of the protocol. We finally ask whether it carries standalone value, that is, whether the artifact, separated from the procedure that produced it, improves performance on unseen tasks.

\textbf{Setup.}
We test whether knowledge distilled by our framework can be separated from the tasks that produced it and reused on unseen tasks. The \emph{donor} is the LLM that produced the knowledge: its agents solved the self-improvement split, and its forums and distillation produced the frozen knowledge asset. The \emph{recipient} is the LLM that consumes this frozen asset on held-out tasks. For each transfer benchmark, we hold out 20 unseen difficult tasks as an evaluation split: from a candidate pool of tasks disjoint from the split used for self-improvement, we run a preliminary seed-0 no-knowledge baseline of each recipient LLM (Haiku 4.5 and GPT-5.4-mini) and keep only tasks that \emph{both} recipients fail, sampling 20 at random from that intersection (Appendix~\ref{app:task-maps}). Every evaluation task is therefore hard for both recipients, and we ask whether the transferred knowledge helps solve them.

We first run the full 10-generation framework on the original self-improvement split (50 tasks), where task solving, discussion, and distillation produce a structured knowledge asset containing cross-task heuristics, constraints, pitfalls, and validation strategies. We then freeze that generation-10 asset and transfer it to a disjoint evaluation split. The recipient run is zero-shot and task-execution-only: it uses no new forum discussion and no recipient-side distillation. A task-conditioned adapter converts the shared donor asset into a short memo tailored to the current task. We observed that the amount of generalizable knowledge varies across benchmarks. To address this, we relaxed the constraints and allowed the agent to dynamically determine how much knowledge to transfer to the target new task, preventing the memory from becoming overly noisy (\cref{app:knowledge-transfer}). This directly probes whether the distilled artifact itself contains portable problem-solving value. We repeat every transfer and no-knowledge condition over three seeds and report mean{\scriptsize$\pm$}standard deviation.

\textbf{Results.}
Table~\ref{tab:kt_heatmap} shows that transferred knowledge improves zero-shot performance on both Polyglot and ARC-AGI-1 in every donor--recipient pairing. Across all cells the GPT-authored bundle is the stronger donor, yet cross-family transfer remains positive in both directions. Because the recipient runs no new forum or distillation, these gains come from the frozen donor bundle at inference time, indicating that the bundle carries donor-agnostic structure rather than donor-specific habits. The Haiku-donor-to-GPT cell on ARC-AGI-1 carries the largest seed variance ({\scriptsize$\pm$}12.6), so we read cross-family magnitudes as indicative rather than precise.

\begin{table}[H]
\centering
\small
\renewcommand{\arraystretch}{1.15}
\setlength{\tabcolsep}{5pt}

\begin{subtable}[t]{0.48\linewidth}
\centering
\textbf{Coding (Polyglot) Task}\\[0.35em]
\begin{tabular}{lccc}
\toprule
& \multicolumn{3}{c}{Knowledge Donor} \\
\cmidrule(lr){2-4}
Recipient & N/A & GPT & Haiku \\
\midrule
GPT
& \ktcell{8}{8.3\%{\scriptsize$\pm$2.9}}
& \ktcell{20}{\textbf{20.0\%{\scriptsize$\pm$5.0}}}
& \ktcell{12}{11.7\%{\scriptsize$\pm$2.9}} \\
Haiku
& \ktcell{3}{3.3\%{\scriptsize$\pm$2.9}}
& \ktcell{12}{11.7\%{\scriptsize$\pm$2.9}}
& \ktcell{12}{11.7\%{\scriptsize$\pm$2.9}} \\
\bottomrule
\end{tabular}
\end{subtable}
\hfill
\begin{subtable}[t]{0.48\linewidth}
\centering
\textbf{Abstract Reasoning (ARC-AGI-1) Task}\\[0.35em]
\begin{tabular}{lccc}
\toprule
& \multicolumn{3}{c}{Knowledge Donor} \\
\cmidrule(lr){2-4}
Recipient & N/A & GPT & Haiku \\
\midrule
GPT
& \ktcell{23}{23.3\%{\scriptsize$\pm$2.9}}
& \ktcell{43}{\textbf{43.3\%{\scriptsize$\pm$2.9}}}
& \ktcell{38}{38.3\%{\scriptsize$\pm$12.6}} \\
Haiku
& \ktcell{13}{13.3\%{\scriptsize$\pm$2.9}}
& \ktcell{28}{28.3\%{\scriptsize$\pm$2.9}}
& \ktcell{23}{23.3\%{\scriptsize$\pm$2.9}} \\
\bottomrule
\end{tabular}
\end{subtable}
\vspace{1em}
\caption{Knowledge Transfer Result. Zero-shot solve rate (\%) on 20 held-out tasks when the curated knowledge artifact (i.e., the knowledge assets from our framework collected over the 10 generations shown in Table~\ref{tab:baseline_reasoning_coding}) is reused on the test set. Darker shading indicates higher solve rate.}
\label{tab:kt_heatmap}
\end{table}

\section{Conclusion}\label{sec:conclusion}

This work establishes a shift from agent-centric to knowledge-centric self-improvement, where the persistent substrate of AI progress is a shared, curated knowledge base rather than the internal parameters or memories of the agents themselves. By instantiating this paradigm through a three-stage knowledge curation protocol of task-level forums, cross-task forums, and distillation, we have demonstrated that even stateless, intentionally simple agents can outperform agent-centric self-improvement baselines. This suggests that the bottleneck in autonomous reasoning may not be the complexity of the agent architecture, but rather the quality and structure of the information it is able to access and refine over time.

Beyond performance gains, the broader implication of this paradigm is the creation of a collaborative and inspectable agent ecosystem. Because the knowledge lives outside any single LLM, it points toward ``plug-and-play'' frameworks where different LLMs and agent variants can contribute to a unified body of reusable understanding. This externalization ensures that improvements are not merely LLM-specific behaviors prone to overfitting, but are instead generalized principles that remain effective across different LLM families and unseen problem domains.


Our experiments deliberately held agent architectures simple to isolate the effects of knowledge curation, and this establishes a foundation for investigating the synergy between knowledge-centric and agent-centric techniques. Integrating the curation protocol with recursive search or complex planning modules is a promising frontier for further performance scaling. The framework's success on the benchmarks reported here also provides a roadmap for scaling toward long-horizon tasks that require subtask decomposition. In these more complex settings, the curation of knowledge between sub-components may become a critical unit of study, potentially revealing deeper hierarchical patterns in how AI systems organize and apply learned logic. Extending the protocol to incorporate human-expert contributions, which we did not study in this paper, is another natural direction.


\section*{Acknowledgments}\label{sec:acknowledgments}
This work was supported in part by NSF \#2505096, \#2240110, and gifts from OpenAI and Point72.


\clearpage


\bibliography{refs}
\bibliographystyle{abbrvnat}


\clearpage
\appendix
\section*{Appendix Contents}
\noindent\begin{minipage}{\linewidth}
\setlength{\parskip}{0.35em}
A.~~\hyperref[app:attempt-runtime]{Attempt Runtime Lifecycle}\dotfill\pageref*{app:attempt-runtime}\par
B.~~\hyperref[app:knowledge-examples]{Curated Knowledge Examples}\dotfill\pageref*{app:knowledge-examples}\par
C.~~\hyperref[app:disagreement-examples]{Disagreement as Evidence: Worked Examples}\dotfill\pageref*{app:disagreement-examples}\par
D.~~\hyperref[app:prompts]{Prompts by Benchmark}\dotfill\pageref*{app:prompts}\par
E.~~\hyperref[app:protocol-schemas]{Protocol Schemas}\dotfill\pageref*{app:protocol-schemas}\par
F.~~\hyperref[app:tool-surfaces]{Tool Surfaces by Benchmark}\dotfill\pageref*{app:tool-surfaces}\par
G.~~\hyperref[app:cost-accounting]{Cost Accounting}\dotfill\pageref*{app:cost-accounting}\par
H.~~\hyperref[app:baseline-provenance]{Baseline Provenance and Deviations from Upstream}\dotfill\pageref*{app:baseline-provenance}\par
I.~~\hyperref[app:hyperparameters]{Hyperparameters}\dotfill\pageref*{app:hyperparameters}\par
J.~~\hyperref[app:task-maps]{Task Maps}\dotfill\pageref*{app:task-maps}\par
K.~~\hyperref[app:benchmark-details]{Benchmark Details}\dotfill\pageref*{app:benchmark-details}\par
L.~~\hyperref[app:knowledge-transfer]{Knowledge Transfer}\dotfill\pageref*{app:knowledge-transfer}\par
\end{minipage}
\bigskip

\section{Attempt Runtime Lifecycle}
\label{app:attempt-runtime}

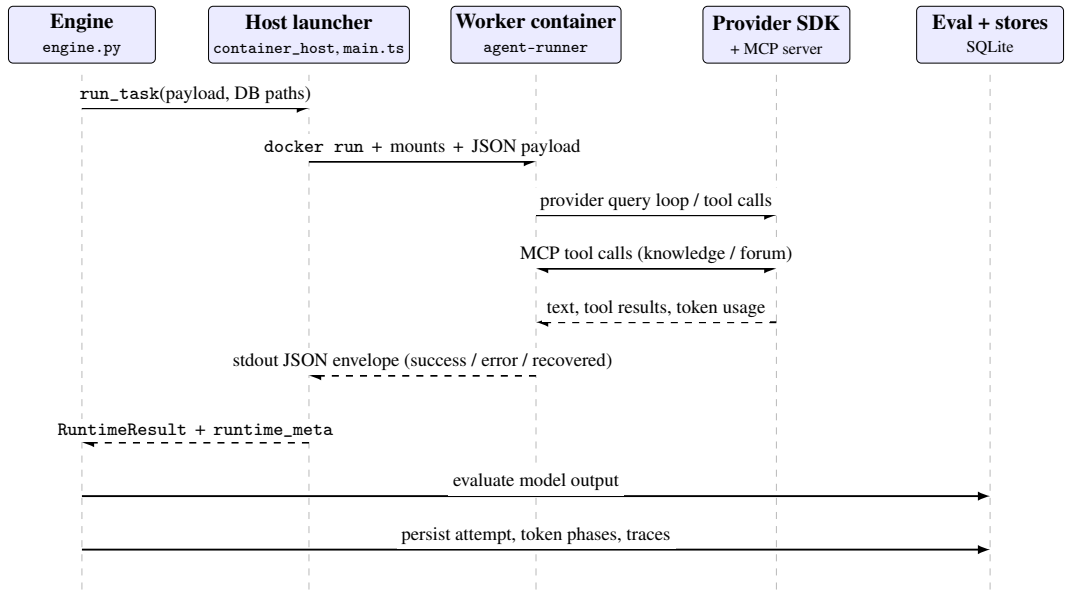
\begin{figure}[H]
\centering
\resizebox{\linewidth}{!}{%
\begin{tikzpicture}[
  font=\small,
  lane/.style={draw, rounded corners=2pt, fill=blue!8,
               minimum height=0.85cm, minimum width=2.2cm,
               align=center, inner sep=2pt, font=\small\bfseries},
  msg/.style={-latex, thick},
  ret/.style={-latex, thick, dashed},
  bidir/.style={latex-latex, thick},
  mlabel/.style={font=\footnotesize, fill=white, inner sep=2pt}
]
\def\xE{0}     \def\xH{3.4}  \def\xC{6.8}  \def\xP{10.4}  \def\xS{13.6}

\node[lane] (E) at (\xE, 0) {Engine\\{\scriptsize\mdseries\texttt{engine.py}}};
\node[lane] (H) at (\xH, 0) {Host launcher\\{\scriptsize\mdseries\texttt{container\_host}, \texttt{main.ts}}};
\node[lane] (C) at (\xC, 0) {Worker container\\{\scriptsize\mdseries\texttt{agent-runner}}};
\node[lane] (P) at (\xP, 0) {Provider SDK\\{\scriptsize\mdseries+ MCP server}};
\node[lane] (S) at (\xS, 0) {Eval + stores\\{\scriptsize\mdseries SQLite}};

\foreach \x in {\xE,\xH,\xC,\xP,\xS}
  \draw[dashed, gray!55] (\x,-0.6) -- (\x,-8.4);

\draw[msg] (\xE,-1.1) -- (\xH,-1.1)
  node[mlabel, midway, above] {\texttt{run\_task}(payload, DB paths)};

\draw[msg] (\xH,-1.9) -- (\xC,-1.9)
  node[mlabel, midway, above] {\texttt{docker run} \,+\, mounts \,+\, JSON payload};

\draw[msg] (\xC,-2.7) -- (\xP,-2.7)
  node[mlabel, midway, above] {provider query loop / tool calls};

\draw[bidir] (\xC,-3.5) -- (\xP,-3.5)
  node[mlabel, midway, above] {MCP tool calls (knowledge / forum)};

\draw[ret] (\xP,-4.3) -- (\xC,-4.3)
  node[mlabel, midway, above] {text, tool results, token usage};

\draw[ret] (\xC,-5.1) -- (\xH,-5.1)
  node[mlabel, midway, above] {stdout JSON envelope (success / error / recovered)};

\draw[ret] (\xH,-6.1) -- (\xE,-6.1)
  node[mlabel, midway, above] {\texttt{RuntimeResult} \,+\, \texttt{runtime\_meta}};

\draw[msg] (\xE,-6.9) -- (\xS,-6.9)
  node[mlabel, midway, above] {evaluate model output};

\draw[msg] (\xE,-7.7) -- (\xS,-7.7)
  node[mlabel, midway, above] {persist attempt, token phases, traces};
\end{tikzpicture}%
}
\caption{Simplified attempt runtime lifecycle. Solid arrows are forward calls, dashed arrows are returns, and the bidirectional arrow denotes in-loop MCP tool exchange. Between the container envelope and the \texttt{RuntimeResult}, the Host parses the envelope and normalizes outcomes (e.g., reclassifying empty success-like envelopes as \texttt{silent\_failure}) so that they are not scored as false zeros.}
\label{fig:attempt-runtime-lifecycle}
\end{figure}

Figure~\ref{fig:attempt-runtime-lifecycle} shows a simplified view of how a single task attempt flows through the runtime. The Engine dispatches the task to a Host launcher, which spawns a transient Worker container; the agent-runner inside the container drives a provider query loop with MCP-served tools, and the JSON status envelope returned through the Host is normalized before the Engine scores the output and persists results.

\section{Curated Knowledge Examples}
\label{app:knowledge-examples}

\begin{figure}[p]
    \centering
    \includegraphics[width=1.0\linewidth]{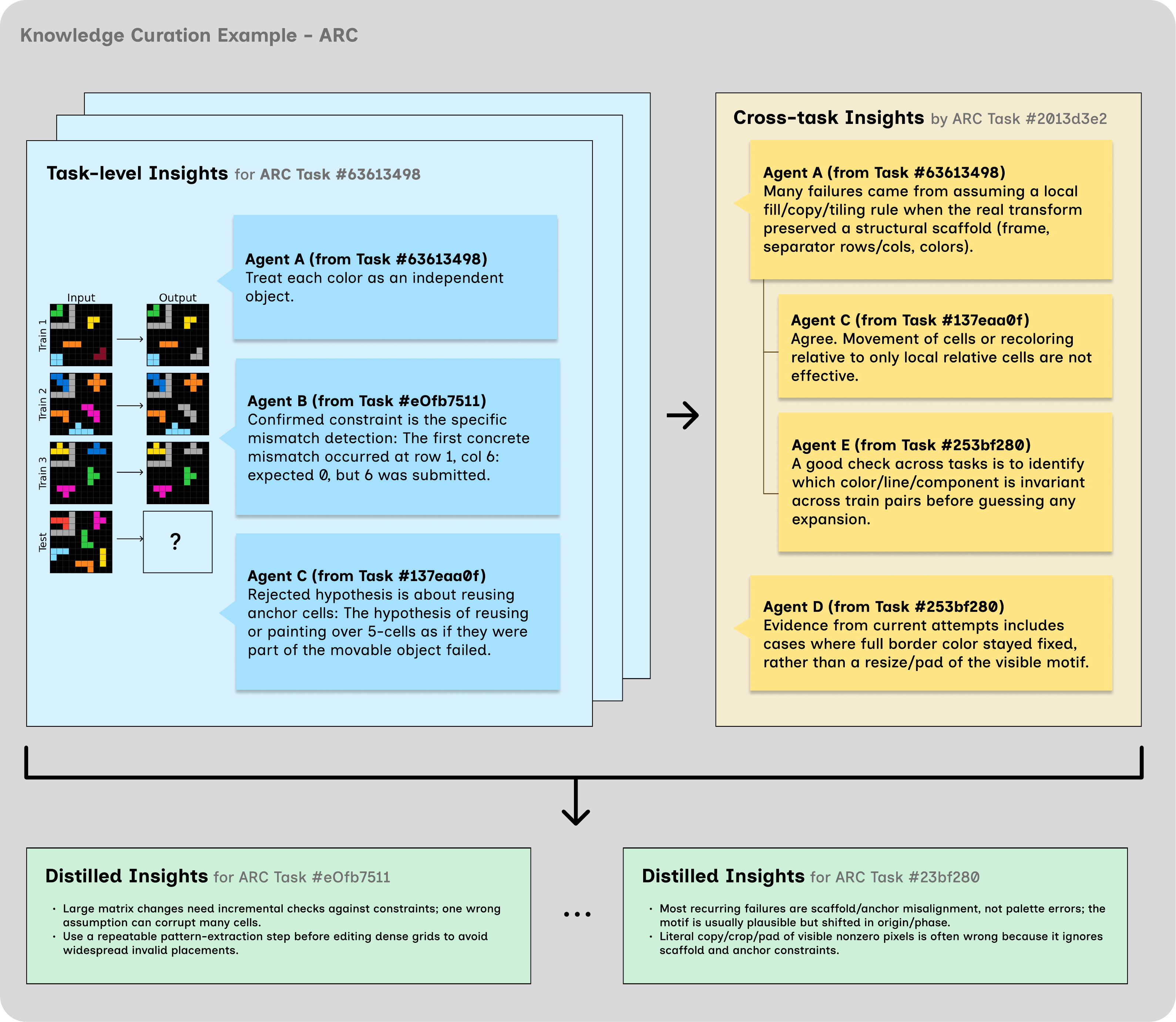}
    \caption{In-depth overview and example of knowledge curation: full curation framework and actual knowledge asset for an ARC task at each step.}
    \label{fig:arc_full_knowledge_ex}
\end{figure}
\clearpage

\begin{figure}[p]
    \centering
    \includegraphics[width=1.0\linewidth]{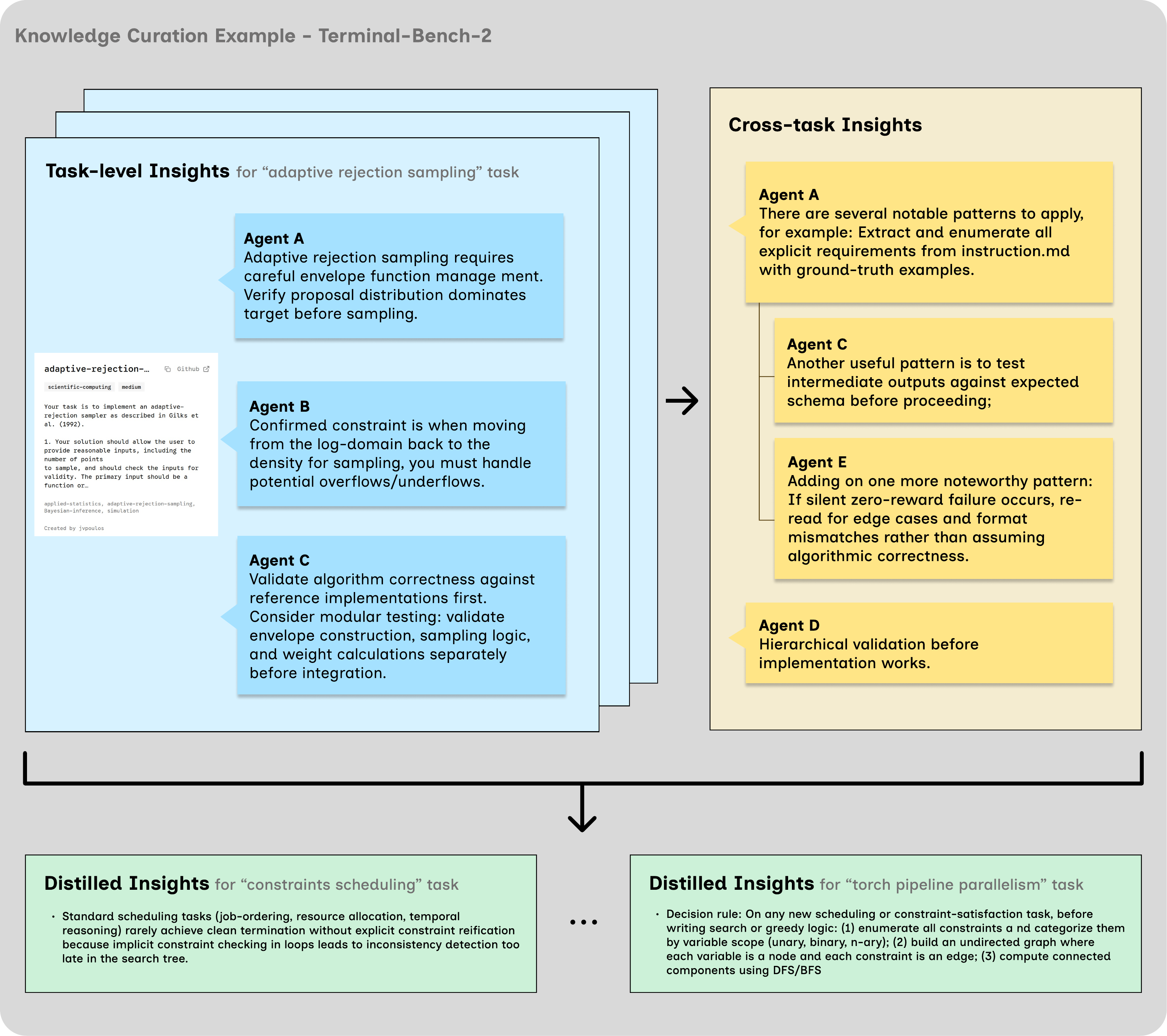}
    \caption{In-depth overview and example of knowledge curation: full curation framework and actual knowledge asset for a Terminal-Bench 2 task at each step.}
    \label{fig:tb2_full_knowledge_ex}
\end{figure}
\clearpage

\section{Disagreement as Evidence: Worked Examples}
\label{app:disagreement-examples}

Section~\ref{sec:methodology} claims that the curation protocol treats disagreement between agents as useful evidence rather than instability. This appendix substantiates that claim with material recorded in the knowledge stores of the main-text self-improvement runs: how often explicit stances occur, and worked examples in which a disagreement changed what was distilled and what a later generation did. Post ids refer to knowledge-entry ids in the corresponding run's store; quotes are abridged (elisions marked \texttt{[...]}, punctuation normalized to ASCII) but otherwise verbatim.

\subsection{Stance Frequencies}
\label{app:stance-frequencies}

Table~\ref{tab:stance-freq} counts explicit stances in cross-task forum discussion rounds ($\geq$1), where the protocol requires each post's \texttt{transfer\_claim} to take an \texttt{AGREE}, \texttt{DISAGREE}, or \texttt{SYNTHESIZE} stance toward a cited peer post (Appendix~\ref{app:protocol-schemas}). Counts are literal keyword occurrences in the posted claim; a post may combine stances (e.g.\ ``DISAGREE + SYNTHESIZE''), so columns are not mutually exclusive. The GPT-5.4-mini runs receive the same prompts but express contrast in free text without emitting the literal keywords, so keyword counts are restricted to the Haiku~4.5 runs.
\begin{table}[h]
\centering
\small
\begin{tabular}{lrrrr}
\toprule
Benchmark & Posts (round $\geq$1) & \texttt{AGREE} & \texttt{DISAGREE} & \texttt{SYNTHESIZE} \\
\midrule
ARC-AGI-1        & 141 & 37  & 10 & 104 \\
ARC-AGI-2        & 123 & 38  & 9  & 81  \\
Polyglot         & 249 & 80  & 14 & 167 \\
SWE-bench Pro    & 265 & 61  & 15 & 201 \\
Terminal-Bench 2 & 576 & 296 & 25 & 308 \\
\midrule
Total            & 1{,}354 & 512 & 73 & 861 \\
\bottomrule
\end{tabular}
\caption{Explicit stances in cross-task forum discussion rounds for the Haiku~4.5 self-improvement runs (seed 1). Stance keywords may co-occur in one post, so columns are not mutually exclusive.}
\label{tab:stance-freq}
\end{table}

Outright disagreement is rare relative to agreement and synthesis (73 of 1{,}354 posts), which is the intended regime: the stance requirement makes conflict explicit and forces it to carry evidence; it does not manufacture conflict.

\subsection{Example 1: Narrowing a Falsified Hypothesis Instead of Discarding the Family (ARC-AGI-1)}
\label{app:disagreement-example-arc}

In generation 1 of the ARC-AGI-1 run (Haiku~4.5, seed 1), the agent on task \texttt{b548a754} assumed that a marker cell (value 8) positioned above a rectangle meant the rectangle stays unchanged, and scored 0.0. Its round-0 cross-task post generalized the failure into a blanket claim:

\begin{nonumberedlisting}
\begin{lstlisting}[style=promptcode,caption={Round-0 overgeneralization (post 196, grounded in the failed \texttt{b548a754} attempt).},label={lst:disagree-arc-r0}]
"concrete_primitive": "Marker (value 8) presence does NOT imply the marker
serves as extension target or directional constraint; must construct
per-pair feature matrix [...] before hypothesizing marker-controlled
geometry."
\end{lstlisting}
\end{nonumberedlisting}

A round-1 post disagreed, citing a solved task in the same generation as a counterexample:

\begin{nonumberedlisting}
\begin{lstlisting}[style=promptcode,caption={Round-1 disagreement (post 240, grounded in solved task \texttt{05f2a901}).},label={lst:disagree-arc-r1}]
"transfer_claim": "DISAGREE with post 196 [...]: task 05f2a901 (score 1.0)
proves marker position DOES directly determine movement direction via
center-point alignment. [...] the transfer claim must be refined --
markers CAN control geometry IF center-point comparison is applied; the
failure in b548a754 was likely assuming extension rather than testing
center-based movement direction."
\end{lstlisting}
\end{nonumberedlisting}

The distiller resolved the conflict by narrowing the falsification to the parameterization that was actually tested, keeping the hypothesis family alive as explicitly untried alternatives:

\begin{nonumberedlisting}
\begin{lstlisting}[style=promptcode,caption={Generation-1 per-task bundle for \texttt{b548a754} (abridged).},label={lst:disagree-arc-bundle}]
"rejected_hypotheses": "FALSIFIED: marker-controlled rectangle extension
  [...] with parameterization 'marker positioned above rectangle ->
  rectangle remains unchanged' (score 0.0) -- UNTRIED: marker selects
  which rectangle transforms, marker position encodes direction/magnitude,
  marker triggers rotation/reflection, marker erases/moves independently"
"next_steps": "Systematically test four hypotheses in order: (1) marker
  moves or disappears between input and output, (2) marker selects which
  rectangle (if multiple exist) undergoes transformation, (3) rectangle
  transforms based on a fixed rule independent of marker position,
  (4) marker position encodes direction or magnitude of rectangle
  transformation."
\end{lstlisting}
\end{nonumberedlisting}

In generation 2 a fresh agent, seeded with this bundle, solved \texttt{b548a754} (score 1.0) with a rule from the kept family: ``The rectangle extends to encompass the marker position in the direction indicated by the marker's location: if marker is below the rectangle $\rightarrow$ extend downward to the marker's row [...]''. Had the swarm adopted the round-0 post as consensus, the marker-direction family would have been marked dead; the disagreement confined the falsification to the one parameterization that had actually failed.

\subsection{Example 2: Splitting One Claim Into Two Scoped Claims (Polyglot)}
\label{app:disagreement-example-polyglot}

In generation 1 of the Polyglot run (Haiku~4.5, seed 1), both \texttt{java\_\_react} and \texttt{rust\_\_react} failed. The \texttt{java\_\_react} agent's round-0 cross-task post (post 157) claimed both languages need the same guard: ``Both java\_\_react and rust\_\_react require guarding callback invocation and downstream propagation with value-change detection''. In round 1 the same agent revised its own claim after reading the Rust attempt's per-task post (post 124), which documented that accumulating changed cells during the propagation loop fires callbacks with intermediate values:

\begin{nonumberedlisting}
\begin{lstlisting}[style=promptcode,caption={Round-1 self-revision (post 213), disagreeing with the author's own round-0 claim.},label={lst:disagree-poly-r1}]
"transfer_claim": "DISAGREE with post 157's claim that java__react and
rust__react both need the same equals() gate. Java's synchronous immediate
propagation [...] reaches stable state in a single call stack, so
previousValue.equals(newValue) suffices. Rust's loop-based propagation
[...] requires a different guard: snapshot before the loop, compare to
final after the loop."
\end{lstlisting}
\end{nonumberedlisting}

Distillation turned the disagreement into a conditional claim plus a falsification, rather than picking a winner:

\begin{nonumberedlisting}
\begin{lstlisting}[style=promptcode,caption={Generation-1 cross-task bundle items citing the disagreement (abridged).},label={lst:disagree-poly-bundle}]
"transferable_insights": "Implement snapshot-before-loop vs. in-place
  comparison based on propagation model: synchronous immediate propagation
  uses in-place equals() checks during recalculation; loop-based
  propagation requires pre/post-loop snapshots to distinguish intermediate
  changes from final stable state."
"rejected_hypotheses": "FALSIFIED: Accumulating changed cells during
  propagation loop iterations and invoking callbacks for all accumulated
  cells after the loop exits, with evidence from rust__react post 213
  [...] UNTRIED: Snapshot-before-loop approach with post-loop comparison"
\end{lstlisting}
\end{nonumberedlisting}

\texttt{rust\_\_react} was solved in generation 2 with a solution whose propagation algorithm implements exactly the scoped guard (``Snapshot all compute cell values before propagation starts [...] Only invoke callbacks for cells where final\_value != snapshot\_value''), and \texttt{java\_\_react} in generation 3.

\subsection{Unresolved Disagreements Are Preserved, Not Forced to Consensus}
\label{app:disagreement-unresolved}

Not every disagreement resolves. On the Polyglot task \texttt{go\_\_connect} (hexagonal-board connectivity, unsolved in all 10 generations of seed 1), the forum debated across generations whether row-parity-dependent neighbor offsets or a uniform 8-neighbor union yields correct hexagonal adjacency; a generation-4 post disagreed with a same-generation proposal by citing generation-3 distilled evidence that parity offsets ``created asymmetric adjacency that caused BFS to miss valid paths''. Neither position produced a solve, and the store retains both as \texttt{FALSIFIED}/\texttt{UNTRIED} markers with their supporting and contradicting evidence rather than a forced consensus. This is the intended failure mode: when the evidence is genuinely conflicting, the protocol's job is to keep the conflict legible to future agents, not to average it away.

\section{Prompts by Benchmark}
\label{app:prompts}

We use a stable prompt scaffold across benchmarks and specialize only the task-facing interface. Each run provides an agent with a benchmark-specific execution prompt and, when applicable, a generated \texttt{TASK.md} plus a prior-generation distilled knowledge asset. After each task execution, our knowledge curation framework invokes agents that write structured post-mortems, forum insights, and cross-task abstractions that can be reused by later generations. 

\subsection{Shared Prompt Scaffold}

Across benchmarks, the execution prompt is procedural: start from the task specification, use memory when available, avoid repeated failed strategies, work toward the required verification or submission step, and return the final answer in the required format. The same reflection layer is reused after task execution. Agents report a load-bearing assumption, evidence, a proposed change for the next generation, a predicted outcome, and confidence. Forum prompts further require concise, evidence-grounded \texttt{INSIGHT} and \texttt{COMMENT} blocks rather than generic advice.

\begin{nonumberedlisting}
\begin{lstlisting}[style=promptcode,language=Python,caption={Illustrative excerpt of the shared execution and reflection scaffold of our framework; abstracted from the \texttt{prompts/} and \texttt{forum/prompt.py} modules.},label={lst:shared-scaffold}]
prompt_lines = [
    "Read the task specification before acting.",
    "Review prior-generation memory when it is available.",
    "Avoid repeating failed approaches recorded in memory.",
    "Work toward the benchmark's required verification or submission step.",
    "Return the final answer in the exact required format.",
]

postmortem_fields = [
    "load_bearing_assumption",
    "evidence",
    "evidence_post_id",
    "proposed_change_for_next_gen",
    "predicted_outcome",
    "confidence",
]

forum_prompt = [
    "Write concise INSIGHT blocks grounded in task evidence.",
    "Write COMMENT blocks only when adding or challenging evidence.",
    "Name concrete primitives; avoid generic advice.",
]
\end{lstlisting}
\end{nonumberedlisting}

\subsection{Benchmark-Specific Adaptation}

The benchmark-specific prompts differ mainly in the task surface and output contract. ARC tasks expose grids and require blind submissions, SWE-bench Pro tasks require repository edits and benchmark verification, Polyglot tasks require complete source-file solutions, and Terminal-Bench 2 tasks require operational success inside a live native container.

\paragraph{ARC.}
\texttt{arc1} and \texttt{arc2} share the same prompt design and are both treated as \texttt{task\_source: arc}; the split and task map distinguish the benchmarks. The workspace is pre-populated with the task data (\texttt{payload.json}, \texttt{grid\_summary.md}, \texttt{TASK.md}) and pre-stubbed attempt files (two files, \texttt{attempt\_1.txt} and \texttt{attempt\_2.txt}, for a single-test task, or one file per (test, trial) pair for a multi-test task); the agent reads the inlined grids directly and submits by overwriting these files with plain ASCII grids. Each attempt file is pre-stubbed with a non-parseable sentinel (\texttt{\_\_NOT\_SUBMITTED\_\_}) that the grid parser rejects, so a run that never overwrites is scored as a no-submission rather than a false zero, and the submission rate equals the actual overwrite rate.

\begin{nonumberedlisting}
\begin{lstlisting}[style=promptcode,language=Python,caption={Illustrative ARC prompt excerpt; abstracted from \texttt{\_build\_arc\_no\_mcp\_execution\_prompt} in \texttt{prompts/\_\_init\_\_.py}.},label={lst:arc-prompt}]
arc_prompt = """\
You are solving one ARC visual reasoning task.

Workspace: /workspace/task/workspace
Public task files (already written for you):
- payload.json       canonical JSON with `train` pairs and `test` inputs
- grid_summary.md    readable rendering of the same grids in plain ASCII
- TASK.md            the same grids inlined in markdown
- attempt_1.txt, attempt_2.txt   placeholder ASCII grids you must overwrite

Goal: infer the train-input -> train-output transformation, apply it to the
test input, and write your answer as plain ASCII grids -- NOT JSON.

Output format (overwrite both files):
  /workspace/task/workspace/attempt_1.txt
  /workspace/task/workspace/attempt_2.txt
Each file is rows of space-separated integers (0-9), one row per line,
matching the format of the `Output:` blocks in grid_summary.md.

Tools: you have Read, Edit, Write, Bash, Glob, Grep. Prefer Read for
inspection; use Bash for short commands and overwriting the attempt files.

Workflow:
  1. Read each training pair; note dimensions, palette, objects, and what
     changed input -> output.
  2. Within your first few turns, write a first-pass guess to BOTH
     attempt_1.txt and attempt_2.txt (a copy of the test input is fine as
     a placeholder) so the workspace always contains a submission.
  3. Form one transformation rule that explains every training pair and
     verify it pair-by-pair before applying.
  4. Apply the rule to the test input and overwrite attempt_1.txt and
     attempt_2.txt with the refined answer (submit two structurally
     different hypotheses if you can construct them; otherwise submit the
     same grid into both files).
  5. Validate before exit:
     python3 /workspace/task/workspace/validate_prediction.py \\
       /workspace/task/workspace/attempt_1.txt \\
       /workspace/task/workspace/attempt_2.txt
"""
\end{lstlisting}
\end{nonumberedlisting}

\paragraph{SWE-bench Pro.}
For \texttt{swebench\_pro}, the prompt frames each task as repository repair. Agents read the issue, write a short checklist of behavior changes, inspect verification targets before editing, identify the implementation path, and make the smallest diff that satisfies the checklist. In upstream-strict mode, \texttt{TASK.md} withholds exact test names and gives count-based verification guidance. In seeded mode, it can expose failing tests, selected test files, and a benchmark wrapper command.

\begin{nonumberedlisting}
\begin{lstlisting}[style=promptcode,language=Python,caption={Illustrative SWE-bench Pro execution and task excerpts; abstracted from \texttt{\_build\_swebench\_execution\_prompt} and \texttt{\_build\_swebench\_task\_markdown} in \texttt{prompts/\_\_init\_\_.py}.},label={lst:swebench-prompt}]
swebench_prompt_lines = [
    "Read `TASK.md` in the active task workspace.",
    "Write a short checklist of required behavior changes before editing.",
    "Inspect the verification section and target tests before editing.",
    "Identify the implementation path and interfaces that must change.",
    "Use regression targets and selected tests from TASK.md when available.",
    "When TASK.md provides a benchmark test-script command, run it "
    "after each edit and iterate on the first failure.",
    "Edit files in place. The runtime captures the workspace diff "
    "as the canonical patch.",
]

swebench_task_lines = [
    "# TASK",
    "",
    f"- task_id: {task.id}",
    f"- task_source: {task_source}",
    f"- instance_id: {instance_id}",
    f"- repo: {repo}",
    "",
    "## Objective",
    "Complete the behavior described by the issue and requirements.",
    "",
    "## Verification",
    f"There are {fail_to_pass_count} target test(s) that must change "
    "from failing to passing.",
    "Benchmark test script command:",
    f"- `{script_command}`",
    "",
    "## Issue",
    task.prompt.strip() or "(none)",
    "",
    "## Output Format (required)",
    "Edit files in the workspace repo. The runtime captures the "
    "workspace diff as the canonical patch.",
]
\end{lstlisting}
\end{nonumberedlisting}

\paragraph{Polyglot.}
For \texttt{polyglot}, the prompt removes the repository-repair framing and treats each task as a language-specific programming exercise. Agents read the statement, inspect tests when the contract is underspecified, implement the required functions or types, handle stated edge cases, and run the real verification command when available. Unlike SWE-bench Pro, the final answer is the complete solution source file or files rather than a captured repository diff.

\begin{nonumberedlisting}
\begin{lstlisting}[style=promptcode,language=Python,caption={Illustrative Polyglot execution and task excerpts; abstracted from \texttt{\_build\_polyglot\_execution\_prompt} and \texttt{\_build\_polyglot\_task\_markdown} in \texttt{prompts/\_\_init\_\_.py}.},label={lst:polyglot-prompt}]
polyglot_prompt_lines = [
    "Read `TASK.md` in the active task workspace.",
    "Read the full problem statement carefully before editing.",
    "Do not expect hidden benchmark tests in the workspace; treat the problem "
    "statement, starter code, and any inlined examples as the specification.",
    "Implement the required functions, methods, or types in the starter files.",
    "Handle edge cases mentioned in the problem statement.",
    "Run the verification command from TASK.md before finalizing when available.",
    "Do not rely on ad hoc snippets as final verification when a real "
    "test command exists.",
    "Return the answer in the exact format specified in TASK.md.",
]

polyglot_task_lines = [
    "# TASK",
    "",
    f"- task_id: {task.id}",
    "- task_source: polyglot",
    f"- language: {language}",
    f"- exercise: {exercise_name}",
    "",
    "## Problem",
    task.prompt.strip() or "(none)",
    "",
    "## Required Verification",
    "The hidden test suite is NOT seeded into the workspace (for most",
    "exercises the tests double as the answer key), so no test file exists",
    "on disk to inspect. Run the real exercise test command below as a",
    "compile/type-check pass when available; do not create, stub, or guess",
    "the missing test file:",
    "```",
    test_command,
    "```",
    "",
    "## Starter Code",
    f"### `{filename}`",
    f"```{ext}",
    content.rstrip(),
    "```",
    "",
    "## Output Format (required)",
    "Return the full source code of the solution file(s), not a diff.",
]
\end{lstlisting}
\end{nonumberedlisting}

\paragraph{Terminal-Bench 2.}
For \texttt{terminal\_bench\_2}, the prompt relies on the benchmark's native task specification rather than a generated \texttt{TASK.md}. Agents are told to read \texttt{tb2/instruction.md} as the authoritative task statement and \texttt{tb2/task.toml} for timeouts and environment. 

\begin{nonumberedlisting}
\begin{lstlisting}[style=promptcode,language=Python,caption={Illustrative Terminal-Bench 2 execution prompt excerpt; abstracted from \texttt{\_build\_terminal\_bench\_2\_execution\_prompt} in \texttt{prompts/\_\_init\_\_.py}.},label={lst:tb2-prompt}]
tb2_prompt_lines = [
    "Read the native `tb2/instruction.md` first. It is the "
    "authoritative task statement.",
    "Read `MEMORY.md` next and actively reuse it before planning or "
    "repeating exploration.",
    "Read `tb2/task.toml` for task metadata, timeouts, and environment hints.",
    "Treat the mounted framework workspace as task specification only; "
    "the real work happens in the native task container filesystem.",
    "After reading the spec files, identify the real task surface quickly: "
    "current directory, repo/app paths, services, ports, config files, "
    "and build/runtime entrypoints.",
    "Prefer short mutate-then-check cycles over long speculative shell blocks.",
    "If the task depends on a service, daemon, port, or deployed artifact, "
    "verify it from a fresh command after reload or restart.",
    "Make the smallest set of changes needed to satisfy the native task "
    "instruction.",
]
\end{lstlisting}
\end{nonumberedlisting}

\subsection{Summary}

The main design choice is to keep our framework's scaffold stable while adapting the task-facing prompt to each benchmark's action surface. ARC prompts emphasize visible grid comparison and blind trial submission. SWE-bench Pro prompts emphasize issue-to-code alignment, targeted tests, and in-place repository patches. Polyglot prompts emphasize direct exercise completion, starter-code editing, and full-source output. Terminal-Bench 2 prompts emphasize native benchmark instructions, live-container state, and operational verifier behavior. This separation lets our framework reuse memory, reflection, and discussion prompts across benchmarks while preserving the verification semantics required by each environment.

\section{Protocol Schemas}
\label{app:protocol-schemas}

This section gives the canonical schemas of the artifacts the curation protocol writes into the knowledge base: per-task forum posts, cross-task forum posts, and the per-task and cross-task distilled bundles seeded into next-generation agents. We also describe the cache-stability split applied to forum-prompt bodies.

\subsection{Per-Task Forum Post Schema}

The body of every \texttt{forum\_post(text=...)} call in a per-task forum round is a single JSON object with exactly these six fields. Vague values (e.g.\ ``read more carefully'') are flagged for rejection by the distiller; the protocol-sanctioned signal for an exhausted search space is the literal string \texttt{EXHAUSTED} in \texttt{proposed\_change\_for\_next\_gen}.

\begin{nonumberedlisting}
\begin{lstlisting}[style=promptcode,language=Python,caption={Per-task forum post schema; canonical source \texttt{forum/prompt.py}.},label={lst:per-task-schema}]
{
  "load_bearing_assumption": "<the ONE assumption your approach relied on. If you failed, name what was wrong with it; if you succeeded, name why it worked. MUST name a specific tool, API, file, data shape, or invariant -- not a framing like 'read more carefully'.>",
  "evidence":                "<a 1-2 sentence quote from your trace, model output, or a cited prior post that supports the assumption.>",
  "evidence_post_id":        <integer id from the prior-posts list, or null if evidence is from your own Phase-1 notes>,
  "proposed_change_for_next_gen": "<ONE concrete change for a next-gen agent on THIS task. Must name a file/tool/API/decision-point and must differ from anything tried before. If the search space is exhausted, set this to 'EXHAUSTED' and explain in evidence.>",
  "predicted_outcome":       "<a falsifiable prediction (e.g. 'tests X and Y will pass; test Z still fails because ...').>",
  "confidence":              "high" | "medium" | "low"
}
\end{lstlisting}
\end{nonumberedlisting}

A retrieval gate enforced by the MCP server (\texttt{ForumProtocolState} in \texttt{memory/mcp\_server.py}) rejects \texttt{forum\_post} unless the agent has already called \texttt{knowledge(task\_id)} \emph{and} \texttt{query(task\_id, query=...)} on that task in the same session. This is what binds posts to retrieved evidence rather than free-form prose.

\subsection{Cross-Task Forum Post Schema}

Cross-task forum posts use a separate sentinel room keyed by \texttt{task\_id="\_\_cross\_task\_\_"} and a different schema. Round 0 seeds the room with concrete primitives observed in each agent's task; rounds $\geq 1$ must take an explicit \texttt{AGREE}, \texttt{DISAGREE}, or \texttt{SYNTHESIZE} stance toward a cited peer post and must ground the claim in a non-empty \texttt{evidence\_task\_ids} list drawn from the Task Evidence Map.

\begin{nonumberedlisting}
\begin{lstlisting}[style=promptcode,language=Python,caption={Cross-task forum post schema; canonical source \texttt{forum/prompt.py}.},label={lst:cross-task-schema}]
{
  "concrete_primitive": "<a single named operation, API call, function/class, error type, file path, language feature, test-runner flag, or numeric invariant -- verbatim from a Phase-1 task or a cited prior post. e.g. 'rust .iter().map(|c| c.to_digit(10)).collect::<Option<Vec<_>>>()' or '|drow| == |dcol| diagonal invariant'. REJECT framings like 'separation of concerns', 'two-phase pipeline', 'pattern', 'approach', 'strategy', 'architecture'.>",
  "task_grounding": {
    "task_id":           "<your Phase-1 task_id, OR a prior-post-cited task_id>",
    "where_it_appeared": "<a verbatim 1-2 sentence quote (>=40 chars) from your task description, reflection, model_output, error message, or cited prior post; must contain a non-stopword from concrete_primitive>",
    "evidence_post_id":  <integer id of the cited prior post, or null if grounding is from your own Phase-1 task>
  },
  "transfer_claim":      "<round 0: name another task in this generation's forum that would benefit, and how. round 1+: MUST cite a peer post by id and use one of AGREE | DISAGREE | SYNTHESIZE.>",
  "anti_meta_self_check":"<one sentence describing why concrete_primitive is NOT lifted-from-tutorial advice. If you cannot answer without re-stating the primitive concretely, the post is meta and is dropped.>",
  "evidence_task_ids":   ["<task_id from the Task Evidence Map that grounds this primitive. round 1+: MUST be a non-empty list; the server rejects a round-1 post that omits it or cites an unknown task id.>", ...]
}
\end{lstlisting}
\end{nonumberedlisting}

The \texttt{anti\_meta\_self\_check} field is part of the schema, not a prompt-level suggestion: posts whose \texttt{concrete\_primitive} cannot be defended against the meta-test are dropped at distillation time. The cross-task retrieval gate requires only \texttt{query(task\_id="\_\_cross\_task\_\_", ...)}.

\subsection{Distillation Bundle Schema}

Distillation produces two bundle types: one per-task bundle for each surviving (un-solved) task, and one cross-task bundle per generation. Bundles share field names but differ in scope and provenance: per-task bundles consume only per-task forum posts for that task; cross-task bundles consume only cross-task forum posts. Their inputs are kept independent so per-task curation cannot be polluted by cross-task speculation, and vice versa (\texttt{distillation/distiller.py}).

\begin{nonumberedlisting}
\begin{lstlisting}[style=promptcode,language=Python,caption={Distillation bundle types; canonical source \texttt{distillation/types.py}.},label={lst:distill-schema}]
@dataclass
class PerTaskBundle:                  # produced by per_task_distill, one per task
    task_id: str
    transferable_insights: list[Insight]
    confirmed_constraints: list[Insight]
    rejected_hypotheses:   list[Insight]
    pitfalls:              list[Insight]
    checks:                list[Insight]
    next_steps:            list[Insight]
    evidence_post_ids:     list[int]   # post ids cited by any item

@dataclass
class CrossTaskBundle:                # produced by cross_task_distill, one per generation
    transferable_insights: list[Insight]
    confirmed_constraints: list[Insight]
    rejected_hypotheses:   list[Insight]
    pitfalls:              list[Insight]
    checks:                list[Insight]
    next_steps:            list[Insight]
    evidence_post_ids:     list[int]

class Insight(TypedDict, total=False):
    text:                "<actionable claim, 'when X, do Y'>"
    applies_when:        "<concrete condition under which the claim holds>"
    does_not_apply_when: "<concrete counterexample / boundary>"
    evidence:            list[Evidence]   # each: {task_id, post_id, quote}
    confidence:          "high" | "medium" | "low"
\end{lstlisting}
\end{nonumberedlisting}

The next-generation seed package is a dictionary with \texttt{per\_task\_bundle} (the bundle for the exact task an agent will attempt) and \texttt{cross\_task\_bundle} (one shared bundle across all tasks in the generation), rendered into \texttt{MEMORY.md} in the agent's workspace via \texttt{runtime/seeding.py}.

\subsection{Cache-Stability Split for Forum Prompts}

Anthropic's prompt cache keys on the content hash up to and including the \texttt{cache\_control} marker, so mutating any byte of the prefix invalidates the cached entry on every call. To keep cache reads non-trivial across forum rounds, every forum-prompt builder returns a \texttt{ForumPromptParts} object with two halves:

\begin{itemize}[leftmargin=1.4em,topsep=0pt,itemsep=2pt]
    \item \texttt{cacheable\_prefix} --- agent- and generation-invariant content: task identifiers, descriptions, MCP tool list, round instructions, the JSON output schema. \texttt{cache\_control} is attached only to a block containing this prefix.
    \item \texttt{variable\_suffix} --- per-agent / per-generation content: this agent's prior attempts, native session memory, and peer posts surfaced in the current round.
\end{itemize}

Adapters with cache-aware delivery (\texttt{anthropic\_direct\_forum}) emit the prefix as a separate \texttt{cache\_control}-marked block and append the suffix as a plain block; adapters that surface the prompt body through a flat file (e.g.\ \texttt{TASK.md}) call \texttt{ForumPromptParts.as\_text()} to recover the concatenated body.

\section{Tool Surfaces by Benchmark}
\label{app:tool-surfaces}

Across all five benchmarks, agents act through the provider's native coding tool surface (file read/edit/write, shell, search) over a prepared workspace, and no benchmark-specific MCP tool API is exposed during task execution. Benchmark-specific logic is handled by workspace setup, trace capture, and post-hoc evaluation rather than a separate task API.

\paragraph{ARC.}
\texttt{arc1} and \texttt{arc2} share the same workspace contract. The difference between them is the dataset and task map, not the tool surface. The agent receives an ARC workspace pre-populated with the task data and pre-stubbed attempt files; web access is disabled, and the agent's submission is captured by overwriting the attempt files. Submissions are blind: the agent receives no per-trial correctness feedback during the run; attempt files left holding the pre-stub sentinel do not parse as grids and are scored as non-submissions, and the scorer parses the attempt files to derive the per-test pass.

\begin{nonumberedlisting}
\begin{lstlisting}[style=promptcode,language=Python,caption={ARC task tool surface and workspace.},label={lst:arc-tool-surface}]
arc_tools = ["Read", "Edit", "Write", "Bash", "Glob", "Grep"]

arc_workspace_files = [
    "payload.json",          # canonical JSON: train pairs + test inputs
    "grid_summary.md",       # ASCII rendering of the same grids
    "TASK.md",               # grids inlined in markdown
    "attempt_1.txt",         # pre-stubbed __NOT_SUBMITTED__ sentinel (overwrite with answer)
    "attempt_2.txt",         # pre-stubbed __NOT_SUBMITTED__ sentinel (overwrite with answer)
    "validate_prediction.py" # local format-check helper
]
\end{lstlisting}
\end{nonumberedlisting}

The agent reads the inlined grids from the workspace files, infers the train-input to train-output transformation, applies it to the test input, and overwrites both \texttt{attempt\_*.txt} files with rows of space-separated digits.

\paragraph{Coding benchmarks.}
Our framework prepares a workspace containing files such as \texttt{TASK.md}, repository or starter-code context, and verification instructions. The agent then uses the selected provider's normal coding tools, such as shell execution, file reading and editing, search/navigation, or patch application. For \texttt{swebench\_pro}, our framework captures the resulting workspace diff as the candidate patch and evaluates it with the SWE-bench Pro harness. For \texttt{polyglot}, our framework evaluates the edited exercise workspace using the language-specific test command and Docker-based evaluator.

\paragraph{Trace accounting.}
All benchmarks write tool events into a unified \texttt{tool\_trace}. The runtime also stores summarized counters in \texttt{runtime\_meta}, including overall \texttt{tool\_call\_counts} and the \texttt{memory\_tool\_call\_counts} / \texttt{forum\_tool\_call\_counts} subsets used in the curation phases.

\section{Cost Accounting}
\label{app:cost-accounting}

Reported costs cover every LLM call made during a run: agent task-execution tokens, per-task reflection, per-task and cross-task forum rounds, and the distillation pass. The HyperAgents and DGM reruns follow the same accounting, including its meta-loop (self-modification) tokens. We report cost in dollars rather than raw token counts: dollar figures use each model's published API rates, including prompt-cache write and read rates, so they reflect what cached and uncached tokens actually cost and remain comparable across methods with different cache profiles.

\section{Baseline Provenance and Deviations from Upstream}
\label{app:baseline-provenance}

Every rerun baseline in Table~\ref{tab:baseline_reasoning_coding} and Table~\ref{tab:baseline_prompt} executes from a fork. No fork alters the baseline's core optimization algorithm. The deviations from upstream fall into five recurring categories: (i) new benchmark adapters, where upstream does not support a benchmark we evaluate; (ii) LLM substitution, so that every method runs on the same LLM; (iii) information-parity gates, on by default, that remove out-of-channel access to hidden tests or gold answers; (iv) network egress isolation; and (v) budget alignment and cost accounting. This appendix documents each fork's deviations per category.

\paragraph{Shared harness alignment (DGM and HyperAgents).}
Both fork-maintained self-improvement baselines receive the same three cross-cutting changes. \emph{Budget parity:} upstream DGM and HyperAgents hard-code a \texttt{600}s per-task agent cap and a \texttt{120}s Polyglot test-execution timeout; the forks instead read the sweep-level per-task timeout on every benchmark path (\texttt{3600}s in the frozen sweeps, twice the \texttt{1800}s cap of our own main runs; Appendix~\ref{app:hyperparameters}) and unify Polyglot test execution at \texttt{180}s, identical to our evaluator, and sampling temperature is pinned to \texttt{0.0} for both runners. \emph{Egress isolation:} task-agent and meta-agent/self-improvement containers are attached to an internal Docker network whose only route out is an allowlisting HTTP CONNECT proxy permitting the provider API (plus PyPI where the loop installs dependencies); this mirrors our runtime's egress containment and closes the channel by which an agent with unrestricted shell access could fetch public copies of hidden tests or solutions. \emph{Cost accounting:} per-call LLM usage records are aggregated into each run's report, including meta-loop and self-improvement tokens, so reported baseline token and dollar figures cover the full loop rather than task execution alone.

\paragraph{DGM.}
Pinned from a fork of upstream \texttt{jennyzzt/dgm}. Upstream ships SWE-bench (Verified) and Polyglot support only; the fork adds a SWE-bench Pro runner that evaluates through the pinned official SWE-bench Pro evaluator, so DGM's SWE-bench Pro cells run an adapter we authored around upstream's unchanged self-improvement loop. The LLM is routed through the shared provider profile (Haiku~4.5 in the reported runs) instead of the DGM paper's Claude~3.5~Sonnet and \texttt{o3-mini}, so our DGM cells are LLM-substituted and not comparable to DGM's published numbers. Information-parity gates, on by default with environment flags that restore the upstream behavior: the self-improvement/diagnosis LLM sees the scalar solved/unsolved outcome rather than upstream's gold answer patches and hidden tests; hidden-test seeding of SWE-bench Pro workspaces is disabled; the Polyglot workspace's git history is scrubbed so the solver cannot recover hidden tests or the reference solution from \texttt{git log}; and evaluations where zero tests execute count as failures rather than vacuous passes. Upstream's committed initial-run artifacts (\texttt{initial/} logs and predictions, \texttt{initial\_polyglot/}) are dropped and regenerated deterministically at setup.

\paragraph{HyperAgents.}
Pinned from a fork of upstream \texttt{facebookresearch/HyperAgents}. Upstream ships a Polyglot domain adapter (plus several non-benchmark research domains) but has no ARC or SWE-bench Pro domain; both are adapters we authored (\texttt{domains/arc}, \texttt{domains/swebench\_pro}) around upstream's unchanged meta-agent loop. Upstream hard-pins \texttt{claude-3-5-sonnet-20241022} for Polyglot; the fork removes this pin and routes Polyglot through the same provider profile as every other domain, so HyperAgents cells are likewise LLM-substituted. Because Claude LLMs emit a native XML tool-call syntax that upstream's response parser does not recognize, the fork adds a parser for that format --- a functional prerequisite of the LLM substitution, not a capability change. Parity gates, on by default: grader hidden-test artifacts and official evaluation outputs are pruned from the tree copied to the meta-agent, the Polyglot workspace's git history is scrubbed, and vacuous exit-0 passes are guarded, mirroring the DGM gates above.

\paragraph{GEPA.}
Pinned from a fork of upstream \texttt{gepa-ai/gepa}. The fork is purely additive with respect to upstream: it adds our ARC and Polyglot runners and their launch scripts. Our runners carry three default-on parity gates: the ARC reflection LM does not receive the hidden ARC test gold; hidden Polyglot test files are absent from the solver workspace and written in only immediately before grading; and hidden test-runner output tails are withheld from the reflective dataset and redacted from captured trajectories.

\paragraph{OpenEvolve.}
The runtime path is the official PyPI package at the same version. The adapter that drives it is our own in-repo code that instantiates OpenEvolve as a prompt-evolution optimizer over the agent's system prompt, scored by the same runtime executor and evaluators as our system; it does not evolve code as AlphaEvolve~\citep{alphaevolve} does, which is why Table~\ref{tab:baseline_prompt} labels it a prompt optimizer.

\paragraph{Terminal-Bench 2.}
The benchmark source is pinned directly at upstream \texttt{harbor-framework/terminal-bench-2} with no fork, and the leaderboard rows in Table~\ref{tab:baseline_terminal} are reported scores under each system's own harness, not reruns (see the table caption).
\section{Hyperparameters}
\label{app:hyperparameters}

We used a per-task runtime timeout of \texttt{1800}s; the larger frozen baseline sweeps raise this cap to \texttt{3600}s. Terminal-Bench~2 is the exception: there each task's native \texttt{task.toml} agent timeout is the sole wall-time bound, following the official harness contract, so the \texttt{1800}s cap does not apply. Forum and cross-task discussion jobs each use a \texttt{900}s timeout. Benchmark-specific evaluator limits are separate from the agent runtime cap: Polyglot keeps a \texttt{180}s test-execution timeout, while SWE-bench Pro uses a \texttt{3600}s harness timeout in the main runs. Polyglot solving sessions additionally use an in-session retry-with-test-feedback budget of \texttt{tries}$=$\texttt{2}, matching the Aider convention: on a failing attempt, the solver sees the last 50 lines of its own test-runner output and gets one more try within the same session. The DGM and HyperAgents reruns follow their upstream-native feedback loops rather than this convention. ARC additionally enforces a fixed blind submission budget of two trials per test input, following the official ARC protocol.

Beyond timeouts, the main-run defaults for our system are: 10 generations over the fixed 50-task pools (Appendix~\ref{app:task-maps}); one per-task forum round and two cross-task forum rounds per generation; temperature \texttt{0.0} for all direct LLM calls (forum, distillation, and claiming --- task execution runs through the provider's agent runtime); \texttt{medium} reasoning effort for GPT-5.4-mini; at most 25 concurrent task containers; up to 3 retries per task on transient infrastructure failures (timeout or container crash) --- a completed attempt, solved or not, is never retried, and failed attempts' tokens are counted in the reported cost. The frozen baseline sweeps run DGM and HyperAgents at the same 50-task, 10-generation budget with temperature \texttt{0.0} and the unified \texttt{3600}s per-task cap (Appendix~\ref{app:baseline-provenance}); GEPA and OpenEvolve iteration counts under the matched dollar budgets are stated with Table~\ref{tab:baseline_prompt}.

\section{Task Maps}
\label{app:task-maps}

We use fixed task maps for the benchmark families reported in the main text. For \texttt{arc1}, \texttt{arc2}, \texttt{swebench\_pro}, and \texttt{polyglot}, we used 50-task subsets at self-improvement (training) time. For \texttt{terminal\_bench\_2}, we used the full 89 tasks to compare against the actual harness engineering frameworks in the leaderboard. For knowledge-transfer evaluation, we additionally used disjoint 20-task maps for \texttt{arc1} and \texttt{polyglot}, selected for headroom as follows. We first built a candidate pool disjoint from the corresponding self-improvement split: for \texttt{arc1}, 80 tasks sampled uniformly at random from the 400-task public evaluation split (a different split from the equal-sized training set that sources the self-improvement pool, Appendix~\ref{app:benchmark-details}); for \texttt{polyglot}, all 175 tasks outside the 50-task self-improvement subset. We then ran a seed-0 no-knowledge solo baseline of each recipient LLM (GPT-5.4-mini and Haiku 4.5) over the pool, took the intersection of their failure sets (tasks that \emph{both} LLMs failed to solve), and sampled 20 tasks uniformly at random from that intersection. The seed-0 selection runs are separate from the three evaluation seeds reported in Table~\ref{tab:kt_heatmap}, so selection does not bias the transfer-versus-baseline comparison; because every selected task is unsolved by both recipients at selection time, the no-knowledge baseline solve rates are low by construction. These maps share no task identifiers with the corresponding self-improvement (donor) maps. For \texttt{polyglot}, 7 of the 20 evaluation tasks are the same Exercism exercise as a donor task in a different language (e.g.\ \texttt{go\_\_ledger} vs.\ \texttt{java\_\_ledger}); the split is therefore disjoint at the task-identifier level but not fully exercise-disjoint, so same-exercise recall may contribute to the \texttt{polyglot} transfer numbers. The \texttt{arc1} maps are disjoint at the task level and have no analogous same-exercise structure.

\taskmaptitle{ARC-AGI-1 Train/Self-improvement Tasks.}
\begin{nonumberedlisting}
\begin{lstlisting}[style=taskmap]
01. 83302e8f                 26. 25d487eb
02. f76d97a5                 27. ea32f347
03. 8efcae92                 28. 1fad071e
04. 1190e5a7                 29. dc0a314f
05. 57aa92db                 30. 6ecd11f4
06. a85d4709                 31. 9ecd008a
07. a416b8f3                 32. b8825c91
08. 890034e9                 33. 25d8a9c8
09. 694f12f3                 34. 776ffc46
10. a1570a43                 35. 9172f3a0
11. 780d0b14                 36. 6cf79266
12. bdad9b1f                 37. caa06a1f
13. 4938f0c2                 38. d22278a0
14. a740d043                 39. 4612dd53
15. 3428a4f5                 40. b6afb2da
16. 6455b5f5                 41. 94f9d214
17. f35d900a                 42. aabf363d
18. 23b5c85d                 43. 5ad4f10b
19. ce22a75a                 44. 1e0a9b12
20. 5582e5ca                 45. b548a754
21. af902bf9                 46. 05f2a901
22. e48d4e1a                 47. 239be575
23. c8f0f002                 48. e8593010
24. 36d67576                 49. 8731374e
25. 6b9890af                 50. e509e548
\end{lstlisting}
\end{nonumberedlisting}

\taskmaptitle{ARC-AGI-1 Knowledge-Transfer Evaluation Tasks.}
\begin{nonumberedlisting}
\begin{lstlisting}[style=taskmap]
01. 15113be4                 11. 9def23fe
02. 18419cfa                 12. ac2e8ecf
03. 1c56ad9f                 13. af22c60d
04. 4aab4007                 14. b20f7c8b
05. 4b6b68e5                 15. b7cb93ac
06. 626c0bcc                 16. b9630600
07. 64a7c07e                 17. ce039d91
08. 79369cc6                 18. e2092e0c
09. 94be5b80                 19. e7b06bea
10. 96a8c0cd                 20. ecaa0ec1
\end{lstlisting}
\end{nonumberedlisting}

\taskmaptitle{ARC-AGI-2 Train/Self-improvement Tasks.}
\begin{nonumberedlisting}
\begin{lstlisting}[style=taskmap]
01. de493100                 26. 27a28665
02. 694f12f3                 27. c909285e
03. c9680e90                 28. 1efba499
04. e7a25a18                 29. a5f85a15
05. 72a961c9                 30. d43fd935
06. 11dc524f                 31. 4612dd53
07. 484b58aa                 32. ea9794b1
08. fcc82909                 33. 91413438
09. 8be77c9e                 34. b9b7f026
10. 84f2aca1                 35. d5d6de2d
11. 6d75e8bb                 36. 9f8de559
12. ed98d772                 37. e9afcf9a
13. d06dbe63                 38. 29623171
14. db118e2a                 39. 5521c0d9
15. 543a7ed5                 40. 1f876c06
16. fd4b2b02                 41. bf699163
17. 834ec97d                 42. 1a6449f1
18. 63613498                 43. e98196ab
19. 9b365c51                 44. e0fb7511
20. e7dd8335                 45. b548a754
21. ea786f4a                 46. 5ad8a7c0
22. 3d6c6e23                 47. 825aa9e9
23. 896d5239                 48. 973e499e
24. 278e5215                 49. 2013d3e2
25. 5034a0b5                 50. 62ab2642
\end{lstlisting}
\end{nonumberedlisting}

\taskmaptitle{SWE-bench Pro Train/Self-improvement Tasks.}
\begin{nonumberedlisting}
\begin{lstlisting}[style=taskmap]
01. instance_gravitational__teleport-af5e2517de7d18406b614e413aca61c319312171-vee9b09fb20c43af7e520f57e9239bbcf46b7113d
02. instance_internetarchive__openlibrary-1894cb48d6e7fb498295a5d3ed0596f6f603b784-v0f5aece3601a5b4419f7ccec1dbda2071be28ee4
03. instance_NodeBB__NodeBB-f2082d7de85eb62a70819f4f3396dd85626a0c0a-vd59a5728dfc977f44533186ace531248c2917516
04. instance_flipt-io__flipt-db1c3b100e231c62f0c90c2ab037614f20a2a63b
05. instance_navidrome__navidrome-3972616585e82305eaf26aa25697b3f5f3082288
06. instance_internetarchive__openlibrary-d8162c226a9d576f094dc1830c4c1ffd0be2dd17-v76304ecdb3a5954fcf13feb710e8c40fcf24b73c
07. instance_gravitational__teleport-eda668c30d9d3b56d9c69197b120b01013611186
08. instance_future-architect__vuls-86b60e1478e44d28b1aff6b9ac7e95ceb05bc5fc
09. instance_internetarchive__openlibrary-bb152d23c004f3d68986877143bb0f83531fe401-ve8c8d62a2b60610a3c4631f5f23ed866bada9818
10. instance_gravitational__teleport-3ff19cf7c41f396ae468797d3aeb61515517edc9-vee9b09fb20c43af7e520f57e9239bbcf46b7113d
11. instance_protonmail__webclients-715dbd4e6999499cd2a576a532d8214f75189116
12. instance_flipt-io__flipt-5c7037ececb0bead0a8eb56054e224bcd7ac5922
13. instance_navidrome__navidrome-23bebe4e06124becf1000e88472ae71a6ca7de4c
14. instance_element-hq__element-web-18c03daa865d3c5b10e52b669cd50be34c67b2e5-vnan
15. instance_future-architect__vuls-2c84be80b65d022c262956cd26fc79d8bb2f7010
16. instance_element-hq__element-web-27139ca68eb075a4438c18fca184887002a4ffbc-vnan
17. instance_ansible__ansible-a20a52701402a12f91396549df04ac55809f68e9-v1055803c3a812189a1133297f7f5468579283f86
18. instance_qutebrowser__qutebrowser-0833b5f6f140d04200ec91605f88704dd18e2970-v059c6fdc75567943479b23ebca7c07b5e9a7f34c
19. instance_flipt-io__flipt-c188284ff0c094a4ee281afebebd849555ebee59
20. instance_navidrome__navidrome-89b12b34bea5687c70e4de2109fd1e7330bb2ba2
21. instance_tutao__tutanota-d1aa0ecec288bfc800cfb9133b087c4f81ad8b38-vbc0d9ba8f0071fbe982809910959a6ff8884dbbf
22. instance_protonmail__webclients-caf10ba9ab2677761c88522d1ba8ad025779c492
23. instance_element-hq__element-web-459df4583e01e4744a52d45446e34183385442d6-vnan
24. instance_future-architect__vuls-aaea15e516ece43978cf98e09e52080478b1d39f
25. instance_ansible__ansible-b2a289dcbb702003377221e25f62c8a3608f0e89-v173091e2e36d38c978002990795f66cfc0af30ad
26. instance_ansible__ansible-5e369604e1930b1a2e071fecd7ec5276ebd12cb1-v0f01c69f1e2528b935359cfe578530722bca2c59
27. instance_qutebrowser__qutebrowser-ed19d7f58b2664bb310c7cb6b52c5b9a06ea60b2-v059c6fdc75567943479b23ebca7c07b5e9a7f34c
28. instance_future-architect__vuls-f0b3a8b1db98eb1bd32685f1c36c41a99c3452ed
29. instance_internetarchive__openlibrary-acdddc590d0b3688f8f6386f43709049622a6e19-vfa6ff903cb27f336e17654595dd900fa943dcd91
30. instance_protonmail__webclients-0ec14e36ceb01ba45602a563e12352af8171ed39
31. instance_ansible__ansible-b6290e1d156af608bd79118d209a64a051c55001-v390e508d27db7a51eece36bb6d9698b63a5b638a
32. instance_gravitational__teleport-3587cca7840f636489449113969a5066025dd5bf
33. instance_internetarchive__openlibrary-43f9e7e0d56a4f1d487533543c17040a029ac501-v0f5aece3601a5b4419f7ccec1dbda2071be28ee4
34. instance_future-architect__vuls-c11ba27509f733d7d280bdf661cbbe2e7a99df4c
35. instance_protonmail__webclients-dfe5604193d63bfcb91ce60d62db2f805c43bf11
36. instance_qutebrowser__qutebrowser-44e64199ed38003253f0296badd4a447645067b6-v2ef375ac784985212b1805e1d0431dc8f1b3c171
37. instance_flipt-io__flipt-36e62baffae2132f78f9d34dc300a9baa2d7ae0e
38. instance_navidrome__navidrome-f7d4fcdcc1a59d1b4f835519efb402897757e371
39. instance_internetarchive__openlibrary-5fb312632097be7e9ac6ab657964af115224d15d-v0f5aece3601a5b4419f7ccec1dbda2071be28ee4
40. instance_navidrome__navidrome-669c8f4c49a7ef51ac9a53c725097943f67219eb
41. instance_flipt-io__flipt-dbe263961b187e1c5d7fe34c65b000985a2da5a0
42. instance_ansible__ansible-3b823d908e8a5d17674f8c26d337d3114b7493b1-v0f01c69f1e2528b935359cfe578530722bca2c59
43. instance_navidrome__navidrome-e12a14a87d392ac70ee4cc8079e3c3e0103dbcb2
44. instance_NodeBB__NodeBB-445b70deda20201b7d9a68f7224da751b3db728c-v4fbcfae8b15e4ce5d132c408bca69ebb9cf146ed
45. instance_ansible__ansible-a02e22e902a69aeb465f16bf03f7f5a91b2cb828-vba6da65a0f3baefda7a058ebbd0a8dcafb8512f5
46. instance_gravitational__teleport-d873ea4fa67d3132eccba39213c1ca2f52064dcc-vce94f93ad1030e3136852817f2423c1b3ac37bc4
47. instance_tutao__tutanota-f3ffe17af6e8ab007e8d461355057ad237846d9d-vbc0d9ba8f0071fbe982809910959a6ff8884dbbf
48. instance_qutebrowser__qutebrowser-bf045f7ec7c27709ea3ef61cf41a24e8fdd2e7da-v059c6fdc75567943479b23ebca7c07b5e9a7f34c
49. instance_qutebrowser__qutebrowser-1a9e74bfaf9a9db2a510dc14572d33ded6040a57-v2ef375ac784985212b1805e1d0431dc8f1b3c171
50. instance_NodeBB__NodeBB-04998908ba6721d64eba79ae3b65a351dcfbc5b5-vnan
\end{lstlisting}
\end{nonumberedlisting}

\taskmaptitle{Polyglot Train/Self-improvement Tasks.}
\begin{nonumberedlisting}
\begin{lstlisting}[style=taskmap]
01. javascript__queen-attack          26. java__forth
02. rust__wordy                       27. python__dominoes
03. python__dot-dsl                   28. go__word-search
04. java__satellite                   29. javascript__simple-linked-list
05. cpp__diamond                      30. go__counter
06. rust__accumulate                  31. java__react
07. go__error-handling                32. javascript__ocr-numbers
08. cpp__queen-attack                 33. python__scale-generator
09. rust__poker                       34. java__go-counting
10. python__sgf-parsing               35. rust__doubly-linked-list
11. rust__react                       36. python__grade-school
12. java__ledger                      37. javascript__forth
13. go__connect                       38. python__wordy
14. rust__macros                      39. java__mazy-mice
15. javascript__triangle              40. cpp__bank-account
16. java__zipper                      41. python__zipper
17. java__bowling                     42. java__custom-set
18. python__tree-building             43. java__rest-api
19. javascript__say                   44. go__transpose
20. java__wordy                       45. rust__gigasecond
21. python__food-chain                46. rust__say
22. javascript__wordy                 47. go__food-chain
23. python__poker                     48. rust__pig-latin
24. javascript__grade-school          49. go__markdown
25. cpp__gigasecond                   50. go__crypto-square
\end{lstlisting}
\end{nonumberedlisting}

\taskmaptitle{Polyglot Knowledge-Transfer Evaluation Tasks.}
\begin{nonumberedlisting}
\begin{lstlisting}[style=taskmap]
01. cpp__circular-buffer              11. java__protein-translation
02. go__beer-song                     12. java__sgf-parsing
03. go__ledger                        13. java__state-of-tic-tac-toe
04. java__affine-cipher               14. java__twelve-days
05. java__change                      15. javascript__go-counting
06. java__dominoes                    16. javascript__phone-number
07. java__hangman                     17. python__forth
08. java__palindrome-products         18. python__paasio
09. java__pig-latin                   19. python__phone-number
10. java__pov                         20. python__simple-linked-list
\end{lstlisting}
\end{nonumberedlisting}

\section{Benchmark Details}
\label{app:benchmark-details}

We evaluate on five benchmarks spanning coding, terminal skills, and abstract reasoning:
\begin{itemize}[leftmargin=*]
        \item \textbf{Polyglot} \citep{polyglot}: The Polyglot benchmark (Aider Polyglot) is an evaluation framework designed to assess the code generation and editing capabilities of LLMs across six languages: C++, Go, Java, JavaScript, Python, and Rust. Comprising 225 Exercism exercises, the suite prioritizes a model's ability to refactor code and iteratively correct errors using automated test feedback. The benchmark serves as a robust testbed for knowledge curation, as recurring failure modes and language-specific idioms surface consistently across tasks. To ensure direct comparability with existing literature, we utilize the 50-task subset adopted by DGM \citep{dgm} and HyperAgents \citep{hyperagents} for the self-improvement phase to enable fairer comparison. This framework provides a rigorous measure of cross-language generalization and the practical utility of agentic LLMs in realistic software development workflows.
        \item \textbf{SWE-bench Pro} \citep{swebenchpro}: SWE-bench Pro is a benchmark designed to provide a rigorous and realistic evaluation of AI agents for software engineering. The benchmark is significantly more challenging than previous SWE-bench families as it tackles four key challenges: (1) data contamination, (2) limited task diversity, (3) ambiguous or underspecified tasks, and (4) unreliable and irreproducible testing environments. From 731 public set instances, we sampled a 50-task subset uniformly at random with a fixed seed.
        \item \textbf{ARC-AGI-1 \& ARC-AGI-2} \citep{arcagi1, arcagi2}: Two abstract visual reasoning benchmarks that require agents to infer transformation rules from a small number of input-output grid examples and apply the inferred rule to a held-out test input. ARC-AGI-1 draws from 400 training-set puzzles; ARC-AGI-2 provides 1000 puzzles designed to be substantially harder, with more compositional and less pattern-matchable rules. We sample 50 tasks from each benchmark for improving the knowledge base with the official benchmark scoring protocols (i.e., exact grid match). ARC benchmarks are particularly well-suited to studying knowledge curation: the underlying transformation rules are discrete and verifiable, and many rules recur across tasks in varied forms, making it natural for agents to accumulate and refine a reusable library of visual reasoning patterns over iterations.
        \item \textbf{Terminal-Bench 2} \citep{terminalbench2}: A benchmark of 89 tasks set in real terminal environments, designed to test whether agents can complete the kind of work practitioners actually perform at the command line. Tasks are inspired by real workflows and span software engineering and system administration (e.g., building the Linux kernel from source), machine learning and model training, security and cryptography, data science, and web/version-control configuration. Each task ships with its own containerized environment, a human-written reference solution, and a comprehensive verification test suite; an attempt counts as solved only when the final environment state passes those tests. Terminal-Bench 2 complements the coding benchmarks by stressing the full agentic workflow loop: agents must issue shell commands, parse their outputs, and recover from errors, rather than producing a single static patch.
\end{itemize}

\section{Knowledge Transfer}
\label{app:knowledge-transfer}

\paragraph{Task-conditioned adapter.}
To make the frozen asset operational at inference time, a task-conditioned adapter converts the shared donor asset into a short memo tailored to the current task before the solver acts. The adapter is a single LLM call (temperature \texttt{0.0}) that is explicitly forbidden from solving the task: given the current task and the shared distilled prior, it selects only the prior items that genuinely apply and emits a compact JSON memo with a fixed schema (\texttt{relevant\_constraints}, \texttt{relevant\_heuristics}, \texttt{pitfalls\_to\_avoid}, \texttt{checks\_before\_submit}, a short task-conditioned \texttt{candidate\_plan}, and a \texttt{knowledge\_use\_rationale}). Rather than transferring a fixed quantity of knowledge, each list field is bounded at \texttt{0--3} items and the adapter is instructed to return short (possibly empty) lists when the prior is only weakly relevant --- this lets the adapter dynamically determine how much knowledge to carry over per task and prevents the recipient memory from becoming noisy or detrimental. The following excerpts are the adapter prompts for Polyglot and ARC.

\begin{nonumberedlisting}
\begin{lstlisting}[style=promptcode,language=Python,caption={Polyglot knowledge-transfer adapter system prompt; excerpt from \texttt{prompts/kt\_adapter.py}.},label={lst:kt-adapter-polyglot}]
KT_ADAPTER_POLYGLOT_SYSTEM = (
  "You are a knowledge-transfer adapter for Polyglot coding tasks. "
  "You are given a current coding exercise and a shared distilled prior "
  "from earlier Polyglot tasks. Do not solve the task. Your only job is "
  "to select the most relevant prior knowledge for this specific task "
  "and turn it into a short actionable memo for the solver.\n\n"
  "Return JSON only with exactly these keys:\n"
  "{ relevant_constraints, relevant_heuristics, pitfalls_to_avoid,\n"
  "  checks_before_submit, candidate_plan, knowledge_use_rationale }\n\n"
  "Rules:\n"
  "- Select only prior items that genuinely match the current coding task.\n"
  "- Include an item ONLY if it is directly supported by the shared prior "
  "AND clearly applies; do NOT invent task-specific rule hypotheses that "
  "are not grounded in the prior -- deriving the rule is the solver's job.\n"
  "- Strongly prefer prior about the ALGORITHM or approach (how to compute "
  "the result, edge-case handling, failure modes) that transfers.\n"
  "- Generic advice is low value; include it only if it clearly constrains "
  "the current task.\n"
  "- Each list field: 0-3 items.\n"
  "- If prior knowledge is contract-specific, do NOT assert it -- phrase it "
  "as an instruction to VERIFY it from the starter file and visible test "
  "stubs before implementing.\n"
  "- candidate_plan must be short and task-conditioned, not finished code; "
  "it describes the APPROACH only, and its first step must be to read the "
  "starter file and test stubs to pin the exact signature and input "
  "representation.\n"
  "- If the prior is only weakly relevant, return very short lists rather "
  "than forcing matches.\n"
  "- If the current task conflicts with prior knowledge, prefer the task."
)
\end{lstlisting}
\end{nonumberedlisting}

\begin{nonumberedlisting}
\begin{lstlisting}[style=promptcode,language=Python,caption={ARC knowledge-transfer adapter system prompt; excerpt from \texttt{prompts/kt\_adapter.py}.},label={lst:kt-adapter-arc}]
KT_ADAPTER_ARC_SYSTEM = (
  "You are a knowledge-transfer adapter for ARC tasks. You are given a "
  "current ARC task and a shared distilled prior from earlier ARC tasks. "
  "Do not solve the task. Your only job is to select the most relevant "
  "prior knowledge for this specific task and turn it into a short "
  "actionable memo for the solver.\n\n"
  "Return JSON only with exactly these keys:\n"
  "{ relevant_constraints, relevant_heuristics, pitfalls_to_avoid,\n"
  "  checks_before_submit, candidate_plan, knowledge_use_rationale }\n\n"
  "Rules:\n"
  "- Select only prior items that genuinely match the current ARC task.\n"
  "- Each list field: 0-3 items.\n"
  "- candidate_plan must be short and task-conditioned, not a final answer.\n"
  "- If the prior is only weakly relevant, return short lists rather than "
  "forcing matches.\n"
  "- If current task evidence conflicts with prior knowledge, prefer the "
  "current task evidence."
)

# Shared user prompt (both benchmarks): interpolate the current task and prior
user = (
  "Current task:\n{json.dumps(task_payload)}\n\n"
  "Shared distilled prior knowledge:\n{json.dumps(asset_payload)}"
)
\end{lstlisting}
\end{nonumberedlisting}



\end{document}